\documentclass[]{template}

\usepackage[utf8]{inputenc}             
\usepackage[T1]{fontenc}                
\usepackage{url}                        
\usepackage{booktabs}                   
\usepackage{multirow}
\usepackage{colortbl}
\usepackage{multicol}
\usepackage{amsfonts}                   
\usepackage{nicefrac}                   
\usepackage{microtype}                  
\usepackage[dvipsnames]{xcolor}         

\usepackage{latexsym}

\usepackage{graphicx}
\usepackage{float}
\usepackage{subcaption}
\usepackage{wrapfig}
\usepackage{lipsum}

\usepackage{bm}

\usepackage{tabularx} 
\usepackage{ragged2e} 
\newcolumntype{L}{>{\RaggedRight\hangafter=1\hangindent=0em}X}

\usepackage{enumitem}

\usepackage{amsmath}
\usepackage{amssymb}
\usepackage{mathtools}
\usepackage{amsthm}

\setboolean{logo}{true}    

\usepackage[linesnumbered,ruled,vlined]{algorithm2e}

\hypersetup{
    colorlinks=true,
    linkcolor=red,
    citecolor=Cerulean,
    filecolor=magenta,      
    urlcolor=magenta,
}

\usepackage[capitalize,noabbrev]{cleveref}
\crefname{section}{§}{§§}
\Crefname{section}{§}{§§}

\usepackage{calligra}
\DeclareMathAlphabet{\mathcalligra}{T1}{calligra}{m}{n}

\usepackage{pifont}

\theoremstyle{plain}

\theoremstyle{definition}

\theoremstyle{remark}

\renewcommand{\paragraph}[1]{\vspace{1mm}\noindent\textbf{#1}}

\DeclareCaptionLabelFormat{cont}{#1~#2\alph{ContinuedFloat}}
\usepackage[most]{tcolorbox}
\tcbset{
  promptbox/.style={
    top=10pt,
    colback=lightgray!20,
    colframe=Black,
    colbacktitle=NavyBlue,
    enhanced,
    center,
    attach boxed title to top center={yshift=-0.1in,xshift=0.0in},
    boxed title style={boxrule=0pt,colframe=white,},
  }
}
\newtcolorbox{promptbox}[2][]{promptbox, title=#2,#1}
\tcbset{
  takeawaybox/.style={
    top=10pt,
    colback=lightgray!20,
    colframe=Black,
    colbacktitle=BurntOrange,
    enhanced,
    center,
    attach boxed title to top center={yshift=-0.1in,xshift=0.0in},
    boxed title style={boxrule=0pt,colframe=white,},
  }
}
\newtcolorbox{takeawaybox}[2][]{takeawaybox, title=#2,#1}
\tcbset{
  observationbox/.style={
    top=10pt,
    colback=lightgray!20,
    colframe=Black,
    colbacktitle=YellowGreen,
    enhanced,
    center,
    attach boxed title to top center={yshift=-0.1in,xshift=0.0in},
    boxed title style={boxrule=0pt,colframe=white,},
  }
}
\newtcolorbox{observationbox}[2][]{observationbox, title=#2,#1}

\usepackage{xspace}

\newcommand\blfootnote[1]{%
  \begingroup
  \renewcommand\thefootnote{}\footnote{#1}%
  \addtocounter{footnote}{-1}%
  \endgroup
}
\newtcolorbox{AIbox}[2][]{aibox,title=#2,#1}
\newcommand{\tabincell}[2]{\begin{tabular}{@{}#1@{}}#2\end{tabular}}
\newcommand{\thiswork} {OPV{}}

\newcommand{\thisbench} {OPV-Bench{}}
\newcommand\modified[1]{\textcolor{black}{#1}}

\newcommand{\eg}{\textit{e.g.}}
\newcommand{\ie}{\textit{i.e.}}

\usepackage[most]{tcolorbox}
\tcbset{
  aibox/.style={
    width=374.18663pt,
    top=10pt,
    colback=white,
    colframe=black,
    colbacktitle=black,
    enhanced,
    center,
    attach boxed title to top left={yshift=-0.1in,xshift=0.15in},
    boxed title style={boxrule=0pt,colframe=white,},
  }
}

\usepackage{CJK}

\title{OPV: \textbf{O}utcome-based \textbf{P}rocess \textbf{V}erifier for \\ Efficient Long Chain-of-Thought Verification}

\author[1,2$*$]{Zijian Wu}
\author[1,3$*$]{Lingkai Kong}
\author[1${*\dag}$]{Wenwei Zhang}
\author[1]{Songyang Gao}
\author[1]{Yuzhe Gu}
\author[1]{Zhongrui Cai}
\author[1]{Tianyou Ma}
\author[1]{Yuhong Liu}
\author[1]{Zhi Wang}
\author[1]{Runyuan Ma}
\author[1]{Guangyu Wang}
\author[1]{Wei Li}
\author[1]{Conghui He}
\author[1,2]{Dahua Lin}
\author[1${\dag}$]{Kai Chen}

\affil[1]{Shanghai AI Laboratory}
\affil[2]{MMLab, The Chinese University of Hong Kong}
\affil[3]{Shanghai Jiao Tong University}

\begin{abstract}

Large language models (LLMs) have achieved significant progress in solving complex reasoning tasks by Reinforcement Learning with Verifiable Rewards (RLVR).
This advancement is also inseparable from the oversight automated by reliable verifiers. However, current outcome-based verifiers (OVs) are unable to inspect the unreliable intermediate steps in the long reasoning chains of thought (CoTs). Meanwhile, current process-based verifiers (PVs) have difficulties in reliably detecting errors in the complex long CoTs, limited by the scarcity of high-quality annotations due to the prohibitive costs of human annotations.
Therefore, we propose the \textbf{O}utcome-based \textbf{P}rocess \textbf{V}erifier (OPV), which verifies the rationale process of summarized outcomes from long CoTs to achieve both accurate and efficient verification and enable large-scale annotation. 
To empower the proposed verifier, we adopt an iterative active learning framework with expert annotations to progressively improve the verification capability of OPV with fewer annotation costs.
Specifically, in each iteration, the most uncertain cases of the current best OPV are annotated and then subsequently used to train a new OPV through Rejection Fine-Tuning (RFT) and RLVR for the next round.
Extensive experiments demonstrate OPV's superior performance and broad applicability. It achieves new state-of-the-art results on our held-out \textsc{\thisbench}, outperforming much larger open-source models such as Qwen3-Max-Preview with an F1 score of 83.1 compared to 76.3.
Furthermore, OPV effectively detects false positives within synthetic dataset, closely align with expert assessment.
When collaborating with policy models, OPV consistently yields performance gains, \eg, raising the accuracy of DeepSeek-R1-Distill-Qwen-32B from 55.2\% to 73.3\% on AIME2025 as the compute budget scales.

\end{abstract}

\begin{document}

\blfootnote{$\dagger$ Corresponding authors: Wenwei Zhang (zhangwenwei@pjlab.org.cn), Kai Chen (chenkai@pjlab.org.cn)}
\blfootnote{$*$ Equal Contributions}
\maketitle


\section{Introduction}
\label{sec: intro}

Large language models (LLMs) have achieved remarkable performance on challenging reasoning tasks~\citep{openai2024o1, deepseekai2025deepseekr1incentivizingreasoningcapability, yang2025qwen3technicalreport, openai2025gptoss120bgptoss20bmodel}. 
This advancement is largely attributed to the growing use of verifiable oversight. Verifiers are crucial components that not only assign rewards in Reinforcement Learning with Verifiable Rewards (RLVR)~\citep{lambert2024tulu} but also select optimal responses in test-time scaling~\citep{zhang2025tts} and benchmark the capabilities of LLMs~\citep{hendryck2021math, he2024olympiadbench}. As LLMs generate increasingly long and intricate chains of thought (CoTs), the fidelity of these verifiers becomes a critical factor that determines the capability and reliability of LLMs.

Existing verifiers fall into two categories, each with its own limitations. 
Outcome-based verifiers (OVs) assess only the final answer against the ground truth, and overlook the reliability of intermediate steps in long CoTs. 
In contrast, process-based verifiers (PVs) \citep{lightman2023prm800k} examine the entire CoT step-by-step to locate errors. However, they struggle with complex reasoning structures in the input CoT and incur prohibitive costs in both automated verification and expert annotations. Previous works~\citep{wang2023mathshepherd, luo2024omegaprm, zhang2025lessons} resort to coarse heuristics for training and fail to provide accurate correctness verdicts or error locations.
This landscape highlights the need for a more accurate and efficient paradigm for long CoT verification.

\begin{figure}[!t] 
  \centering
  \includegraphics[width=0.85\linewidth]{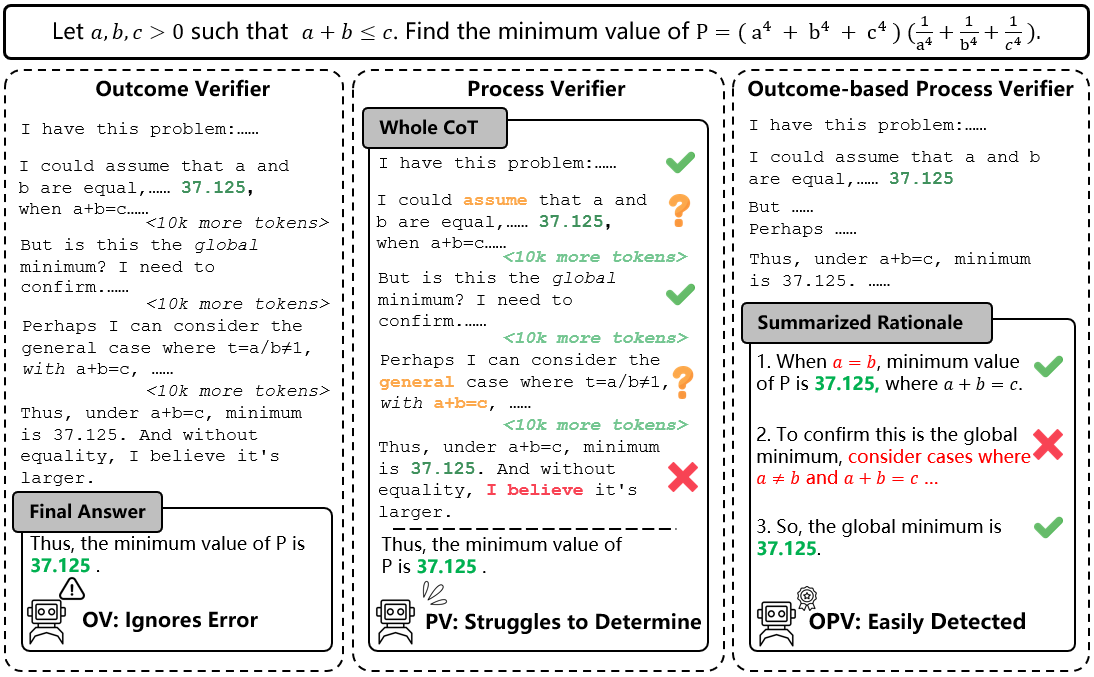}
  \caption{Comparison of different verification paradigms. The policy model happens to guess the correct answer, but fails to solidly prove it, resulting in flawed reasoning. \textbf{Left:} The Outcome-based Verifier ignores underlying reasoning failures. \textbf{Middle:} The Process-based Verifier examines the complex thinking process step-by-step, struggling to identify the tricky logic dependencies. \textbf{Right:} The \textbf{O}utcome-based \textbf{P}rocess \textbf{V}erifier efficiently detects potential process errors from the summarized rationale.
  }
  \label{fig:teaser_comparison}
  \vspace{-12pt}
\end{figure}

To this end, we propose the \textbf{O}utcome-based \textbf{P}rocess \textbf{V}erifier (OPV), a process verifier that operates on summarized outcomes from long CoTs (Fig.~\ref{fig:teaser_comparison}). 
Analogously to summarization, our approach first preserves only the key steps that contribute to the final answer and discards redundant components (\eg, trial-and-error attempts, recalculations, self-overturned assumptions) to form a concise solution path. Then it performs step-by-step verification on the summarized outcome and presents a correctness verdict and, if incorrect, the error location. Compared with OVs, OPV provides more fine-grained supervision, which is useful for the policy model. Moreover, its summarization process significantly reduces the complex, redundant reasoning structures in the input CoT, making OPV more efficient and less susceptible to interference from redundancy than vanilla PV. The simplified CoT also facilitates human annotation, which allows for collecting large-scale, fine-grained expert annotations for training. 

\begin{wrapfigure}{r}{0.43\columnwidth}
  \centering
  \includegraphics[width=\linewidth]{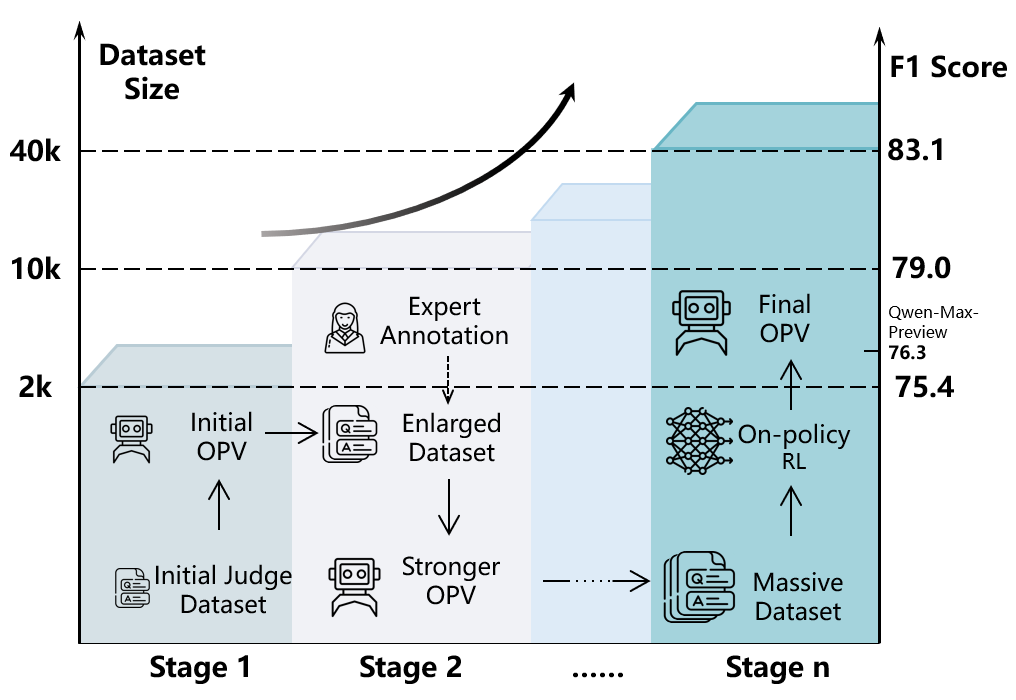}
  \caption{
    Our iterative framework yields enlarged judgment dataset and improved verifier performance.
  }
  \label{fig:scaling_dataset}
\end{wrapfigure}

To streamline massive expert annotation, we adopt an iterative human-in-the-loop framework driven by active learning (Fig.~\ref{fig:scaling_dataset}). 
In each round, the current OPV evaluates each summarized solution multiple times and selects the most uncertain cases for annotation. Expert annotators then provide natural language explanations, correctness verdicts, and error localizations. The newly annotated data are incorporated to retrain the OPV using a combination of off-line rejection fine-tuning and on-line reinforcement learning. 
This strategy effectively strengthens the verifier by focusing on its weaknesses under limited annotation budgets.
After several iterations, we curated 40k annotated solutions across diverse domains of problems spanning from K-12 to undergraduate levels, including a high-quality held-out evaluation set of 2.2k sample answers, namely the \textsc{\thisbench}. 

Empowered by active learning with fine-grained supervision, OPV demonstrates strong performance and broad applicability. Despite its compact size, it achieves performance comparable to much larger open-source models across multiple public and internal benchmarks, including our held-out \textsc{\thisbench}. We further validate OPV's versatility in assisting reasoning models across multiple stages, from training to inference. 
On AM-DeepSeek-R1-0528-Distilled~\citep{ji2025amthinking}, a widely used synthetic dataset verified solely by final answers, OPV identifies false positives at an estimated rate of 7.0\%, closely aligning with expert assessments. In collaboration with various policy models, OPV consistently enhances their test-time performance, with the improvement margin growing as the compute budget scales. For instance, it boosts the accuracy of DeepSeek-R1-Distill-Qwen-32B from 55.2\% to 73.3\% on AIME2025.


\section{Method}

We define the Outcome-based Process Verifier (OPV), a novel framework that bridges outcome and process verification through a faithful, verifiable proxy for long CoT verification (\S~\ref{subsec: task formultaion}). To build a competent OPV model, our approach leverages an iterative active learning framework to select high-quality data for expert annotations (\S~\ref{subsec: active learning}) and a combination of off-line and on-line learning approaches to update the verifier (\S~\ref{sub sec: reinforce learning}). 
We also illustrate the statistics of the annotated data (\S~\ref{subsec:statistics}).

\subsection{Task Formulation}
\label{subsec: task formultaion}



Recent reasoning LLMs generate long chains of thought (CoTs) to solve mathematical problems. These CoTs consist of numerous sequential steps with complex inter-step dependencies, making verification particularly challenging. Two dominant verification paradigms have emerged.

\begin{itemize}
    \item \textbf{Outcome-based Verifier (OV)} checks only the final answer against the ground truth.

    \item \textbf{Process-based Verifier (PV)} sequentially verifies each step throughout the whole CoT.
\end{itemize}
Both approaches have limitations. OV suffers from false positives --- accepting correct answers derived from flawed reasoning — and cannot pinpoint errors in incorrect solutions. PV, with its fine-grained nature, struggles with intricate dependencies of long CoTs.
Moreover, verifying lengthy CoTs is computationally expensive for verifier models and labor-intensive for human annotators. 

To bridge this gap, we propose the \textbf{Outcome-based Process Verifier (OPV)}, a hybrid paradigm that balances faithfulness and efficiency. OPV first summarizes verbose and meandering CoT trajectories into concise, linear solution paths, retaining only the key steps that contribute to the final result while pruning redundant explorations. This summary serves as a faithful proxy of the underlying reasoning rationale, enabling both efficient verification and large-scale human annotation. The verifier then performs a step-by-step validation on this summary to identify the first erroneous step. 
Given a CoT generated for a problem $P$, a summarizer is first applied to produce a structured, $n$-step solution $\mathcal{S} = \{ s_0, \ldots, s_{n - 1}\}$. Subsequently, our OPV, denoted as $\pi$, takes the problem and the structured solution as input and predicts the index $\hat{\ell}$ of the first incorrect step, together with a natural language explanation $\hat{\mathcal{E}}$
\begin{equation}
(\hat{\mathcal{E}}, \hat{\ell}) \sim \pi(\cdot \mid P, \mathcal{S}), \quad \hat{\ell} \in \{-1, 0, \ldots, n-1\}.
\end{equation}
Here, $\hat{\ell} = -1$ indicates a fully correct solution. We focus on identifying the first error, as subsequent steps — though potentially valid in isolation — are built upon faulty premises and thus lack mathematical soundness.

\subsection{Active Learning Framework}
\label{subsec: active learning}
\begin{figure}[t!]
\centering
        \includegraphics[width=0.8\linewidth]{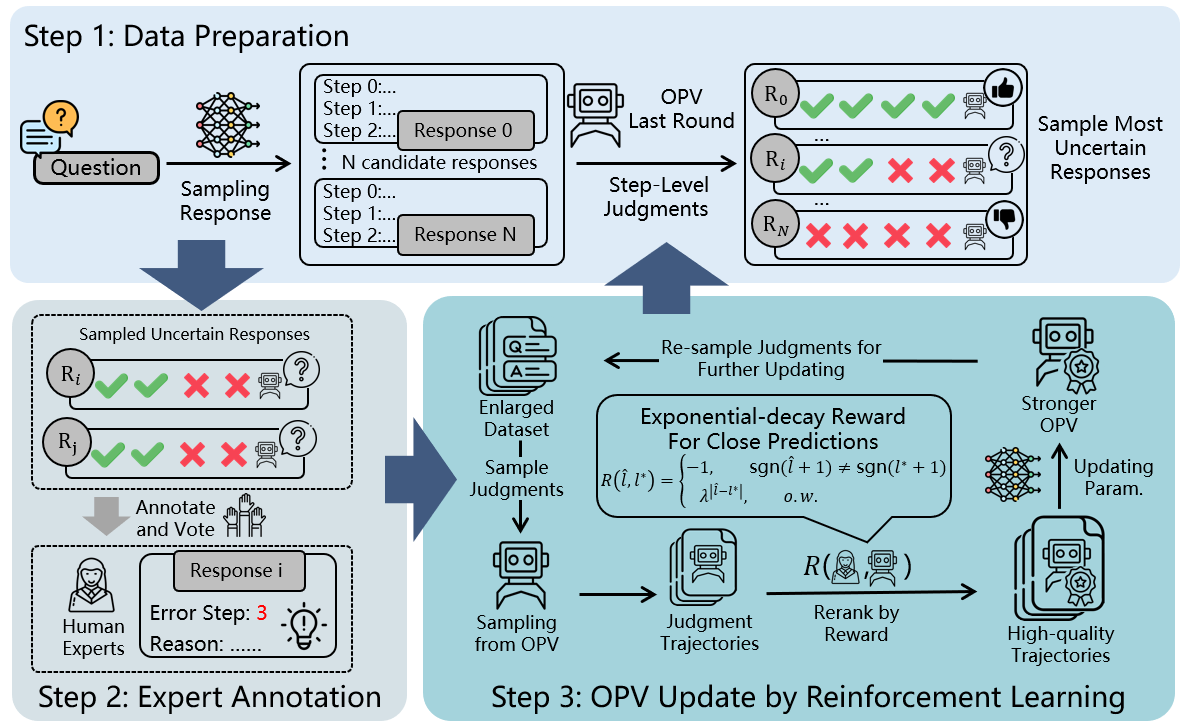}
        \captionof{figure}{\textbf{Overview of our iterative active learning framework.} \textbf{Step 1:} For each sampled solution from the data pool, the verifier generates $N$ candidate judgment trajectories and constructs a training set selected by uncertainty scores. \textbf{Step 2:}  Human experts annotate the most uncertain solutions with ground truth error positions and explanations, expanding the dataset for OPV refinement. \textbf{Step 3:} We then sample high-quality trajectories from the annotated data. The current OPV is then updated using these trajectories with off-line or on-line algorithms.}

        \label{fig:schema}
\end{figure}

Finding potential errors within an answer, even after summarization, is a challenging task. Therefore, it is essential to maximize the utilization of human annotation. To achieve this, we have constructed an iterative human-in-the-loop active learning framework, as shown in Fig.~\ref{fig:schema}. We start with our base verifier $\pi_0$.
In each round, we first use our best OPV model to identify the most uncertain cases for annotation. After human annotation, we then use a combination of off-line expert iteration and on-line reinforcement learning to maximally utilize the information obtained from annotation.

\noindent\textbf{Data Preparation.} 
Initially, we constructed a large data pool of high-quality problems and to-be-verified solutions sampled from top-tier models, represented by the unlabeled data pool $\mathcal{D}_{\mathrm{U}}=\{(P_i, \mathcal{S}_i)\}$. All sampled solutions are pre-summarized by DeepSeek-V3~\citep{deepseekv3} to preserve core rationale. Our set of annotation labels $\mathcal{D_L}_0$ is empty at annotation round $0$. We also established the annotation protocol for our annotators. The complete details are available in Appendix~\ref{subsec: annotation protocol}

\noindent\textbf{Selection Strategy.}
Since verifiers exhibit uneven competency across different types of mathematical reasoning errors, random sampling wastes resources on cases already handled well. We therefore select the most uncertain cases at each round $t$.
For each unlabeled pair $(P_i, \mathcal{S}_i) \in \mathcal{D_U}$, the current OPV $\pi_t$ performs $N$ independent verifications to obtain a set of predicted indices $(\hat{\mathcal{E}}_i^{(j)}, \hat{\ell}_{i}^{(j)})$, 
where $\hat{\mathcal{E}}^{(j)}$ and $\hat{\ell}_{i}^{(j)}$ are the explanation and index predicted by the $j$-th roll-out.

To quantify the model's uncertainty, we compute a consistency score based on the frequency of the most common prediction:
\begin{equation}
\text{Consistency}(\mathcal{S}_i) = \frac{1}{N} \max_{\ell} \sum_{j=1}^N \mathbb{I}\left[\hat{\ell}_{i}^{(j)} = \ell\right].
\end{equation}
A lower consistency score indicates higher uncertainty due to disagreement among the verifier's predictions. We then identify the most uncertain cases based on the consistency score:

\vspace{-1em}
\begin{equation}
\mathcal{Q}_t = \left\{ (P_i, \mathcal{S}_i) \in \mathcal{D_U} \mid \text{Consistency}(\mathcal{S}_i) < \tau_t \right\},
\end{equation}
where $\tau_t \in (0, 1)$ is a dynamically adjustable threshold controlling annotation cost. To mitigate possible overconfidence, we also sample a small proportion of high-consistency data.

\noindent\textbf{Human Annotation.}
We then send the selected data $\mathcal{Q}_t$ for expert annotation. 
Expert annotation yields new labeled data entries with ground truth error positions $\ell_i^*$ and reasons $\mathcal{E}_i^*$ to o enlarge the annotation set
\begin{equation}
\mathcal{D_L}_{t+1} \leftarrow \mathcal{D_L}_t \cup \left\{ (P_i, \mathcal{S}_i, \mathcal{E}_i^*, \ell_i^*) \right\}_{(P_i, \mathcal{S}_i) \in \mathcal{Q}_t}.
\end{equation}
We then train on the new dataset $\mathcal{D_L}_{t+1}$ to obtain a stronger OPV model $\pi_{t+1}$.

\subsection{OPV Update by Reinforcement Learning}
\label{sub sec: reinforce learning}
Human annotation only provides the index of the error with a concise explanation, not a thorough reasoning trajectory that checks each step in detail. Therefore, we use a combination of on-line and off-line approaches to refine the OPV model after each round of annotation. The updated model is then used again to select data for the next round of annotation.

\paragraph{Expert Iteration.}
 For each annotated data entry $(P, \mathcal{S}, \mathcal{E}^*, \ell^*)$, we sample multiple verification attempts from the current OPV. We also sample from other powerful models such as those in the R1 and Qwen families to further boost performance. We retain only those generated verifications that are most consistent with the annotation (\ie, where the predicted index $\hat{\ell}$ matches the ground truth $\ell^*$). These valid verification trajectories are then added to the global verification dataset to update the OPV model. Following expert iteration~\citep{anthony2017thinkingfastslowdeep}, we iterate this process to maximize performance gains.

\paragraph{On-line Reinforcement Learning.} We also use on-line reinforcement learning approaches to stimulate the verification ability.
To ensure stable on-line RL training, we filter the dataset by excluding: (1) highly ambiguous cases that challenge even experts, avoiding annotation noise; and (2) trivial cases with obvious errors, preventing bias toward oversimplified patterns. Then, given a model-predicted index $\hat{\ell}$ and ground-truth index $\ell^*$, we define the exponential-decay reward as follows:
\[
R(\hat{\ell}, \ell^*) =
\begin{cases}
-1 & \text{if } \text{sgn}(\hat{\ell}+1)\neq\text{sgn}({\ell}^*+1), \\
\lambda^{|\hat{\ell}-\ell^*|} & \text{otherwise}.
\end{cases},
\]
where $\lambda \in (0,1)$ controls the penalty for localization errors. The reward is strongly negative only when misclassifying correctness (correct as incorrect or vice versa). Otherwise, it remains positive with exponential decay based on distance error. This design addresses the sparse reward problem for challenging samples requiring precise error localization. We then adopt the DAPO algorithm~\cite{yu2025dapoopensourcellmreinforcement} to obtain our final OPV model.

\subsection{Final Dataset Statistics}
\label{subsec:statistics}
Our framework improves the OPV model's performance while simultaneously establishing a massive, high-quality dataset. After completing all rounds of the "annotate-then-train" process, we progressively scale up the dataset to over 40k expert annotations and over 80k high-quality judge trajectories.
The final fully annotated set spans multiple difficulty levels and knowledge domains.
Particularly, we meticulously curate high-quality samples for evaluation and construct a held-out evaluation set of 2.2k samples, namely the \textsc{\thisbench}, to effectively estimate the verification ability of the trained \thiswork{} model. The detailed breakdown of all annotated data and the \textsc{\thisbench} is available at Appendix~\ref{appendix sec:dataset statistics}.


\section{Experiment}
\label{sec: experiment}

\subsection{Experiment Setup}
\label{subsec:experiment-setup}
\paragraph{Implementation.} In our framework, we use the R1-Distill-Qwen-32B~\citep{deepseekai2025deepseekr1incentivizingreasoningcapability}  model
 to fine-tune the OPV. Further implementation details can be found in Appendix~\ref{appendix sec: implement_detail}.

\paragraph{Evaluation.} We evaluate model performance on \textsc{\thisbench} and \textsc{ProcessBench}~\citep{zheng2024processbench} using three distinct correctness criteria with varying levels of stringency. The \textit{precise} criterion requires exact identification of the erroneous step for a judgment to be considered correct. The \textit{approximate} criterion adopts a more error-tolerant approach, accepting predictions as correct when the identified step is adjacent to the actual error position. Finally, the \textit{rough} criterion considers any error detection for an incorrect answer as a correct judgment. Under each criterion, we compute accuracy separately for erroneous and correct samples, and calculate the harmonic mean of precision and recall of correct samples to obtain the F1 score.

We then compare the performance of \thiswork{} with various state-of-the-art open-source models, including the widely-used Deepseek-R1-0528, Qwen3-Max-Preview, and gpt-oss-120b. We apply the same prompt engineering approach to repurpose these models as critic models. The prompt template is provided in Appendix~\ref{appendix: prompt}. 
To underscore the superior quality of our expert annotations compared to heuristic labeling, we also evaluate Qwen2.5-Math-PRM-72B, a discriminative process reward model trained on labels that integrate Monte Carlo estimation with LLM-as-a-judge.
To better demonstrate the effectiveness of our training framework, we also include several intermediate models (OPV-Stage1 and OPV-Stage2) for comparison. These models have undergone different rounds of annotate-then-train iterations and are trained on different amounts of annotated data, as shown in Fig.~\ref{fig:scaling_dataset}. The final OPV model is obtained after the final stage.

\subsection{Benchmark Evaluation}
\begin{table}[t!]
  \centering
  \vspace{2mm}
  \caption{
  Evaluation results on ProcessBench and \thisbench.
  We report accuracy and F1 scores for three evaluation standards: precisely/absolutely identifying erroneous steps, approximately identifying erroneous steps (within ±1 steps), and roughly identifying whether the whole solution contains errors. All results are reported under \textit{maj@8} voting.
  }
  \vspace{-1mm}
  \scalebox{0.85}{
    \begin{tabular}{lcccccc}
    \toprule
    \textbf{Model} & \tabincell{c}{\textbf{Precise/Abs.} \\ \textbf{Accuracy}} & \tabincell{c}{\textbf{Precise/Abs.} \\ \textbf{F1}} & \tabincell{c}{\textbf{Approximate} \\ \textbf{Accuracy}} & \tabincell{c}{\textbf{Approximate} \\ \textbf{F1}} & \tabincell{c}{\textbf{Rough} \\ \textbf{Accuracy}} & \tabincell{c}{\textbf{Rough} \\ \textbf{F1}} \\
    \midrule
    \multicolumn{7}{c}{\textit{\textbf{ProcessBench (With Standard Answers)}}} \\
    \midrule

Qwen3-Max-Preview & 83.2 & 78.9 & 89.4 & 85.5 & 95.4 & 93.2 \\
DeepSeek-V3-0324 & 74.5 & 71.7 & 83.7 & 79.9 & 94.4 & 92.0 \\
DeepSeek-R1-0528 & 82.2 & 77.7 & 88.7 & 84.6 & 95.4 & 93.1 \\
gpt-oss-120b (high) & 83.3 & 78.8 & 89.2 & 85.1 & 95.7 & 93.5 \\
Qwen2.5-Math-PRM-72B & 77.2 & 74.0 & 84.9 & 81.2 & 93.7 & 91.1 \\
R1-Distill-Qwen-32B & 64.7 & 60.8 & 73.6 & 67.5 & 87.4 & 81.3 \\
OPV-32B & 80.1 & 76.2 & 87.6 & 83.8 & 96.2 & 94.4 \\
    \midrule
    \multicolumn{7}{c}{\textit{\textbf{ProcessBench (Without Standard Answers)}}} \\
    \midrule

Qwen3-Max-Preview & 84.4 & 79.9 & 89.8 & 85.9 & 95.2 & 92.9 \\
DeepSeek-V3-0324 & 72.1 & 69.8 & 81.0 & 77.2 & 91.5 & 88.4 \\
DeepSeek-R1-0528 & 83.2 & 79.3 & 89.3 & 85.8 & 96.0 & 94.1 \\
gpt-oss-120b (high) & 84.7 & 80.3 & 90.4 & 86.6 & 96.2 & 94.2 \\
Qwen2.5-Math-PRM-72B & 76.7 & 73.2 & 83.9 & 79.8 & 91.2 & 87.8 \\
R1-Distill-Qwen-32B & 75.8 & 73.6 & 83.6 & 80.5 & 93.3 & 91.1 \\
OPV-32B & 80.9 & 76.8 & 88.1 & 84.1 & 95.8 & 93.8 \\
    \midrule
    \multicolumn{7}{c}{\textit{\textbf{\thisbench (With Standard Answers)}}} \\
    \midrule

Qwen3-32B & 68.0 & 73.4 & 72.4 & 76.2 & 77.8 & 79.9 \\
QwQ-32B & 65.2 & 71.6 & 70.3 & 74.8 & 75.6 & 78.2 \\
DeepSeek-V3-0324 & 67.9 & 71.7 & 72.9 & 75.0 & 78.6 & 79.1 \\
DeepSeek-R1-0528 & 67.0 & 71.2 & 72.3 & 74.6 & 78.3 & 79.0 \\
Qwen3-Max-Preview & 66.4 & 71.2 & 71.7 & 74.6 & 78.0 & 79.0 \\
gpt-oss-120b (high) & 61.0 & 66.6 & 69.8 & 72.1 & 77.7 & 77.8 \\
Qwen2.5-Math-PRM-72B & 55.4 & 66.5 & 58.4 & 70.0 & 66.0 & 72.3 \\
R1-Distill-Qwen-32B & 70.5 & 76.3 & 74.9 & 79.1 & 78.7 & 81.7 \\
OPV-Stage1 & 68.8 & 75.7 & 72.6 & 78.0 & 76.1 & 80.2 \\
OPV-Stage2 & 71.6 & 75.9 & 76.2 & 79.0 & 81.1 & 82.6 \\
OPV-32B & 78.9 & 79.1 & 83.3 & 82.7 & 87.2 & 86.2 \\
    \midrule
    \multicolumn{7}{c}{\textit{\textbf{\thisbench (Without Standard Answers)}}} \\
    \midrule

Qwen3-32B & 61.2 & 68.7 & 67.0 & 72.1 & 74.2 & 76.7 \\
QwQ-32B & 58.2 & 66.7 & 64.2 & 70.1 & 70.9 & 74.2 \\
DeepSeek-V3-0324 & 60.8 & 67.8 & 66.0 & 70.8 & 72.8 & 75.2 \\
DeepSeek-R1-0528 & 56.9 & 64.7 & 63.2 & 68.2 & 71.7 & 73.6 \\
Qwen3-Max-Preview & 61.0 & 67.3 & 67.0 & 70.8 & 75.2 & 76.3 \\
gpt-oss-120b (high) & 57.9 & 64.1 & 66.8 & 69.3 & 75.6 & 75.4 \\
Qwen2.5-Math-PRM-72B & 55.1 & 66.0 & 58.2 & 67.6 & 63.4 & 70.4 \\
R1-Distill-Qwen-32B & 61.7 & 71.1 & 65.0 & 72.9 & 69.4 & 75.5 \\
OPV-Stage1 & 60.9 & 70.3 & 64.8 & 72.4 & 69.8 & 75.4 \\
OPV-Stage2 & 64.4 & 70.3 & 70.1 & 73.8 & 77.6 & 79.0 \\
OPV-32B & 71.9 & 74.7 & 78.2 & 79.1 & 83.1 & 83.1 \\
    \bottomrule
    \end{tabular}%
    }
  \label{tab:combined_results}%
\end{table}%

Our experiments on two benchmarks reveal distinct challenges in evaluating reasoning verifiers. Tab.~\ref{tab:combined_results} presents results on both \textsc{ProcessBench} and \textsc{\thisbench}, highlighting their different characteristics and difficulty levels.
\textsc{ProcessBench}, which samples answers from LLMs without thinking capabilities, exhibits performance saturation. As shown in the "Rough F1" column, most long reasoning models successfully detect the \textit{existence} of errors in the reasoning process, achieving F1 scores above 90\%. This saturation indicates that error patterns in \textsc{ProcessBench} are more readily identifiable and less representative of sophisticated reasoning failures. The limitation arises because models without thinking mechanisms typically produce more elementary errors, whereas newer thinking LLMs generate more nuanced and subtle logical flaws. Moreover, \textsc{ProcessBench} exclusively comprises problems with explicit, verifiable outcomes, which are more straightforward compared to proof-based problems that demand complex multi-step reasoning.

In contrast, \textsc{\thisbench} presents significantly more complex challenges to verifiers. The test set encompasses a wider spectrum of problems and requires more advanced skills to identify exact errors. Our iterative training paradigm demonstrates its effectiveness by boosting a 32B model's performance above much larger models. Notably, most open-source reasoning models struggle with identifying error positions (see Appendix~\ref{appendix sec: detailed results} for detailed breakdown). While these models achieve high recall, their poor precision indicates an inability to effectively detect errors in solutions—a limitation possibly inherited from training solely on verifiable outcomes.

\section{Application}
\label{sec: application}
We further explore various applications of OPV in this section, demonstrating how it facilitates both the training and inference phases of LLM development.

\subsection{Examining \textsc{AM-DeepSeek-R1-0528-Distilled} using \thiswork}
A primary application of OPV is providing fine-grained supervision for robust training. Outcome-verified synthetic datasets often contain false positives. By using OPV for process verification, we can identify and remove these instances, yielding higher-quality datasets for supervised fine-tuning. 

\begin{figure}[htbp]
\centering
\begin{minipage}{0.48\textwidth}
    \centering
    \includegraphics[width=\textwidth]{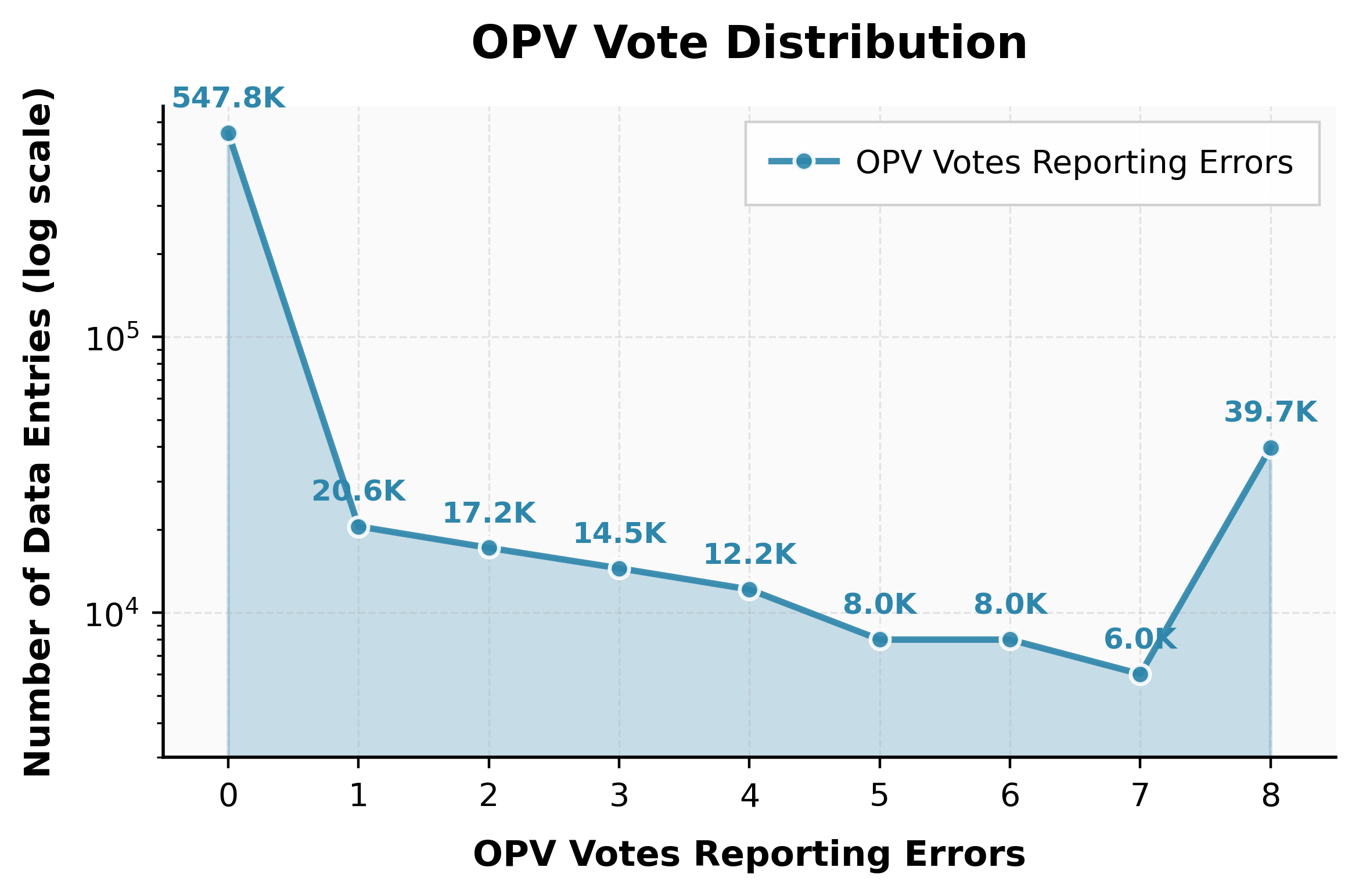}
        \caption{The distribution of \thiswork's votes on AM-DeepSeek-R1-0528-Distilled. 0 means \thiswork{} does not report any error among the 8 votes.}
    \label{fig:opv_amthinking_distribution}
\end{minipage}
\hfill
\begin{minipage}{0.48\textwidth}
    \centering
    \begin{tabular}{lcc}
    \toprule
    Error Category & Count & \% \\
    \midrule
    Logical Process Errors & 24 & 48.0 \\
    Non-Math Problem & 1 & 2.0 \\
    Hallucination & 19 & 38.0 \\
    \quad Problem Modification & \quad 10 & \quad  20.0 \\
    \quad Added False Conditions & \quad 9 & \quad  18.0 \\
    \midrule
    Correct & 6 & 12.0 \\
    \quad Poor Summary & \quad 2 & \quad  4.0 \\
    \quad Fully Correct & \quad 4 & \quad  8.0 \\
    \midrule
    \textbf{Total} & \textbf{50} & \textbf{100.0} \\
    \bottomrule
    \end{tabular}
    \captionof{table}{Error Analysis by Human Experts}
    \label{tab:human_sample_verify}
\end{minipage}

\end{figure}

To validate this approach, we evaluated AM-DeepSeek-R1-0528-Distilled using OPV. Each data entry was verified 8 times, with entries flagged as problematic if OPV reported errors $\ge 6$ times.  We conducted human evaluation on 50 randomly sampled problems to check the reliability of OPV under this setting. Out of 674k math-related data entries checked, 53.7k were flagged as problematic by OPV. The distribution of OPV votes and human evaluation results are available at Fig.~\ref{fig:opv_amthinking_distribution} and Tab.~\ref{tab:human_sample_verify}. Human evaluation results show that OPV demonstrates high reliability in verification, with 88\% of the judgments being valid. Therefore, it is estimated that more than $\frac{53.7}{674.0}\times88\%=7.0\%$ data entries checked contain process errors.

For cost efficiency, we used vanilla summaries rather than re-summarizing the thinking content.
However, some answers might be incorrectly flagged due to inappropriate summarization. We note that 2 of the 50 solutions checked (marked as "Poor Summary") were actually correct but identified as incorrect by \thiswork{} because their original summaries introduced logical gaps. This again highlights the importance of re-summarization for precise verification.

\subsection{Scaling of collaborative reasoning}

Beyond evaluating our OPV on static benchmarks, we study whether it can improve test-time performance in collaboration with policy models. In a collaborative setting, a policy model first samples $N$ complete solutions for a given problem. The verifier then checks each solution $M$ times to estimate its correctness. Finally, we select the answer by aggregating verification verdicts across all solutions.

We conduct experiments on AIME2025 and consider both moderate-sized distilled models and top-tier models as policies. We use OPV as the verifier and define the verification pass rate as the proportion of runs in which the verifier deems the solution correct. We set $N=8$ and $M=16$. When multiple answers tie with the same frequency, we report the average accuracy across the tied answers. Notably, we directly employ the summary part of the original CoT here to reduce compute budget and avoid introducing extra noise. We evaluate the following collaborative strategies and compare them against verifier-free majority voting:

\textbf{Majority Voting.} Among the $N$ sampled solutions, choose the most frequent answer.

\textbf{Best-of-$N$.} Rank solutions by their verification pass rate and output the answer from the top-ranked solution.

\textbf{Verifier Voting.} Use the verification pass rate as the weight of each solution and select the answer with the highest verifier-weighted frequency.

\begin{wrapfigure}{r}{0.5\textwidth} 
  \centering
  \includegraphics[width=0.5\textwidth]{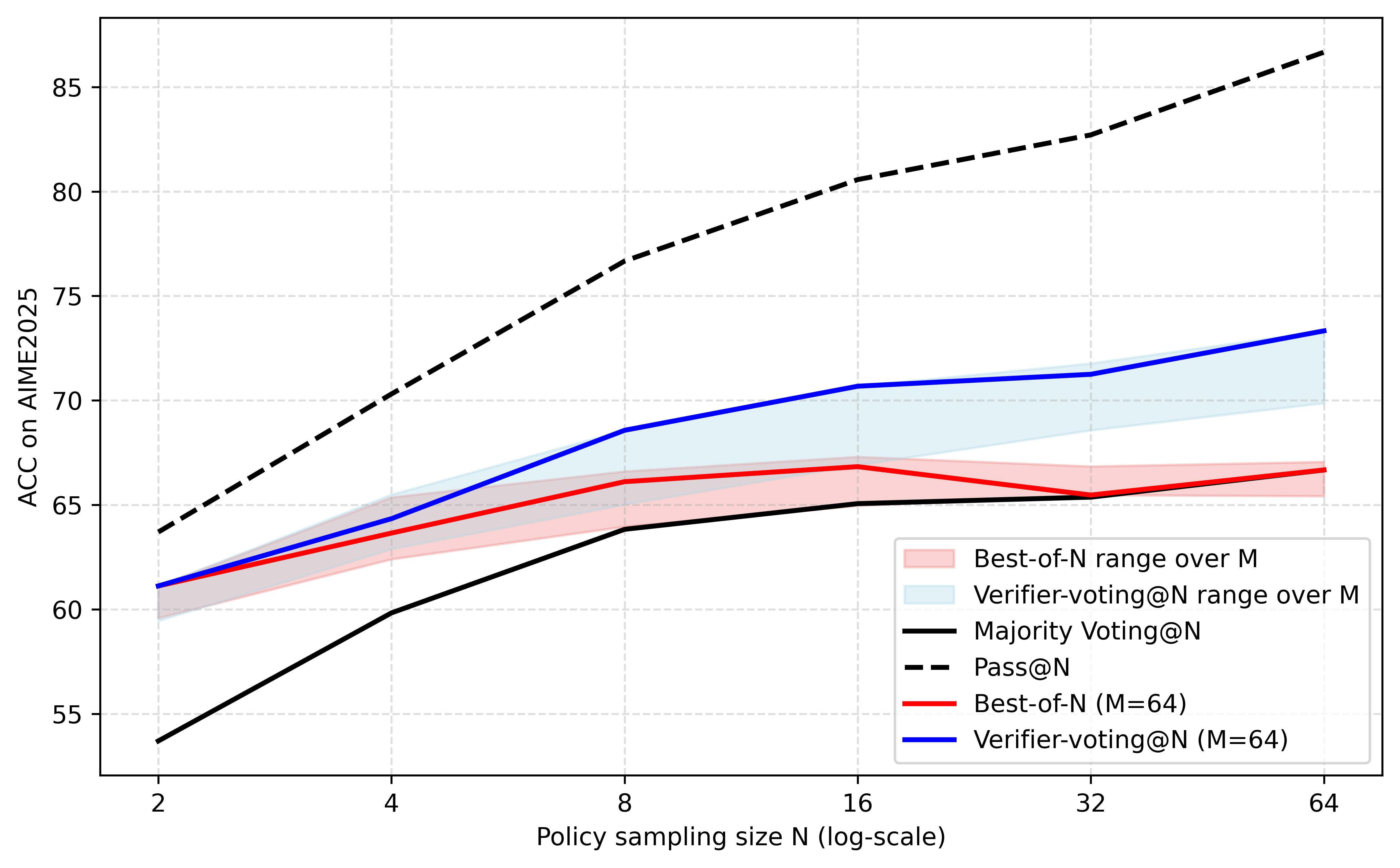}
  \caption{Performance of collaborative reasoning scaling along policy sampling size $N$ and verifier sampling size $M$. Colored bands denote the area covered by accuracy curves as \(M\) increases.}
  \label{fig:aime2025_scaling_curve}
\end{wrapfigure}

As shown in Tab.~\ref{tab:collab-scaling}, OPV consistently boosts performance across all policies in the collaborative setting. For distilled models, verifier voting yields substantial gains of 6.1\% on average over majority voting. Even for top-tier models where majority voting already achieves high accuracy, voting with OPV still provides stable improvements of 2.3\% on average. Notably, verifier voting matches Pass@8 (the 8-sample oracle) for {gpt-oss-120b}. Moreover, verifier voting outperforms Best-of-8 on most policies, indicating that aggregating verification pass rates across multiple candidates is generally more robust than selecting a single best solution.

\begin{table}[t]
\centering
\caption{Performance of different policies and collaborative reasoning strategies on AIME2025. \textbf{Pass@1}: single-sample accuracy averaged over all samples. \textbf{Pass@8}: oracle success with up to eight samples (correct if any of the eight is correct), serving as an upper bound for 8-sample strategies.}
\label{tab:collab-scaling}
\resizebox{\linewidth}{!}
{
\begin{tabular}{lccccc}
\toprule
\textbf{Policy} & \textbf{Pass@1} & \textbf{Majority Voting@8} & \textbf{Best-of-8} & \textbf{Verifier Voting@8} & \textbf{Pass@8} \\
\midrule
DeepSeek-R1-Distill-Qwen-7B & 39.1 & 49.2 & \textbf{56.3} & 55.4 & 66.7 \\
DeepSeek-R1-Distill-Qwen-32B & 55.9 & 63.8 & 66.6 & \textbf{68.0} & 76.7 \\
DeepSeek-R1-Distill-Llama-70B & 44.0 & 50.0 & 56.1 & \textbf{57.8} & 70.0 \\
DeepSeek-R1-0528 & 87.1 & 88.3 & 87.5 & \textbf{90.8} & 96.7 \\
gpt-oss-120b & 92.3 & 93.3 & 95.1 & \textbf{96.7} & 96.7 \\
QwQ-32B & 70.0 & 78.3 & 76.2 & \textbf{80.0} & 83.3 \\
Qwen3-235B-A22B-Thinking-2507 & 95.4 & 96.7 & 97.9 & \textbf{98.3} & 100.0 \\
\bottomrule
\end{tabular}
}
\end{table}

We further evaluate how collaborative reasoning scales with the policy sampling size $N$ and the verifier sampling size $M$. The evaluation focuses on DeepSeek-R1-Distill-Qwen-32B as the policy, scaling N and M from 1 to 64, respectively. Initial sampling uses $N=64$ and $M=64$. For each $(N, M)$ configuration, random subsets are chosen to evaluate different strategies' performance, with 64 repetitions to determine average scores.

Fig.~\ref{fig:aime2025_scaling_curve} demonstrates that accuracy improves with larger N across all strategies, with OPV-enabled Best-of-N and Verifier-Voting consistently outperforming Majority-Voting. Verifier-Voting achieves the highest performance, reaching $73.3\%$ at $N=64$ and $M=64$ — a $6.7$ point improvement over Majority-Voting. Best-of-N gradually converges to Majority-Voting as N increases, indicating that integrating OPV's verification with voting mechanisms yields superior overall performance. These results confirm that OPV collaborates effectively with policy models, with gains increasing proportionally to larger computational budgets. Besides, although the policy model’s native summary part is more concise than the summarized rationale used during training, OPV still proves effective, which can be attributed to the annotation protocol’s tolerance for minor logical gaps.





\section{Related Work}
\label{sec:related_work}

\noindent\textbf{LLM Reasoning.}
Reasoning is regarded as a core capability toward artificial general intelligence. LLMs predominantly perform reasoning in a chain-of-thought (CoT) manner~\citep{wei2022cot}. Frontier LLMs~\citep{openai2024o1, deepseekai2025deepseekr1incentivizingreasoningcapability} further extend the length and complexity of their CoTs to solve challenging math problems.
Rejection Fine-Tuning (RFT)~\citep{yuan2023rft} and Reinforcement Learning with Verifiable Rewards (RLVR)~\citep{lambert2024tulu} rely on verifiers to curate high-quality training data. Recent works like OREAL~\citep{lyu2025oreal} underscore the importance of well-designed verifiers in Reinforcement Learning (RL). 
Motivated by these observations, we develop a verifier that generates verification trajectories in a CoT manner and a framework that iteratively updates it using RFT and RL.

\noindent\textbf{Outcome-based verification vs. process-based verification.}

Outcome-based verifiers (OVs) assess solutions solely by whether the final answer matches the ground truth. Rule-based verifiers, like the Math-Verify library from HuggingFace\footnote{https://github.com/huggingface/Math-Verify}, and LLM-as-a-judge approaches, like CompassVerifier~\citep{liu2025compassverifier}, make simple answer checks scalable. 
However, OVs overlook the reliability of intermediate steps. In contrast, 
process-based verifiers (PVs) meticulously verify the reasoning process step-by-step. 
Empirical evidence shows that PVs can outperform OVs on difficult mathematical benchmarks~\citep{lightman2023prm800k}. 
Yet the prohibitive cost of fine-grained process annotation poses a practical bottleneck. Prior work has therefore resorted to coarse heuristics for training. Monte Carlo methods have been used to assess and assign credit to intermediate steps~\citep{wang2023mathshepherd, luo2024omegaprm}, which can introduce simulation bias and label noise. Another line of work~\citep{zhang2025lessons,yang2025beyondthefirsterror, she2025rprm, duan2025actprm} uses stronger teacher models as judges to provide step-level labels, though the resulting verifiers are ultimately bounded by the teachers' capacity. 
With the rapid progress of LLM reasoning, the possibility of using powerful LLMs to generate verification trajectories in a CoT manner has been explored~\citep{zhang2024genver, mahan2024genrm, shi2025heimdall}. Our OPV follows this generative paradigm but, crucially, engages human experts to provide finer-grained supervision than heuristics, aiming to overcome the inherent limitations of purely model-driven verification.
Similarly, ProcessBench~\citep{zheng2024processbench} also engages human experts for process annotations, while we extend to a more demanding setting with more challenging queries and longer CoTs from frontier LLMs. By summarizing long CoTs into concise rationales and then verifying them, we enable larger-scale expert supervision to power our OPV.


\section{Conclusion}

We introduced the Outcome-based Process Verifier (OPV), which bridges outcome and process verification by operating on summarized solutions from long CoTs. Through an iterative active learning framework with expert annotations, OPV progressively improves its verification capabilities while minimizing annotation costs. Our approach achieves state-of-the-art results across multiple benchmarks, outperforming much larger models including DeepSeek-R1, despite its compact size.

OPV demonstrates broad applicability throughout the reasoning pipeline: it identifies false positives in outcome-verified synthetic data, and yields consistent gains when collaborating with policy models at inference time. The accompanying \textsc{\thisbench} dataset of 2.2k expert-annotated solutions provides a valuable resource for future research.

By enabling efficient and accurate process verification at scale, OPV addresses a critical bottleneck in developing reliable reasoning systems. As LLMs tackle increasingly complex problems with longer reasoning chains, the principle of verifying summarized rationales offers a scalable path toward more trustworthy AI-generated reasoning.

\clearpage
\bibliographystyle{plain}
\bibliography{refs}

@misc{deepseekai2025deepseekr1incentivizingreasoningcapability,
      title={DeepSeek-R1: Incentivizing Reasoning Capability in LLMs via Reinforcement Learning}, 
      author={DeepSeek-AI and Daya Guo and Dejian Yang and Haowei Zhang and Junxiao Song and Ruoyu Zhang and Runxin Xu and Qihao Zhu and Shirong Ma and Peiyi Wang and Xiao Bi and Xiaokang Zhang and Xingkai Yu and Yu Wu and Z. F. Wu and Zhibin Gou and Zhihong Shao and Zhuoshu Li and Ziyi Gao and Aixin Liu and Bing Xue and Bingxuan Wang and Bochao Wu and Bei Feng and Chengda Lu and Chenggang Zhao and Chengqi Deng and Chenyu Zhang and Chong Ruan and Damai Dai and Deli Chen and Dongjie Ji and Erhang Li and Fangyun Lin and Fucong Dai and Fuli Luo and Guangbo Hao and Guanting Chen and Guowei Li and H. Zhang and Han Bao and Hanwei Xu and Haocheng Wang and Honghui Ding and Huajian Xin and Huazuo Gao and Hui Qu and Hui Li and Jianzhong Guo and Jiashi Li and Jiawei Wang and Jingchang Chen and Jingyang Yuan and Junjie Qiu and Junlong Li and J. L. Cai and Jiaqi Ni and Jian Liang and Jin Chen and Kai Dong and Kai Hu and Kaige Gao and Kang Guan and Kexin Huang and Kuai Yu and Lean Wang and Lecong Zhang and Liang Zhao and Litong Wang and Liyue Zhang and Lei Xu and Leyi Xia and Mingchuan Zhang and Minghua Zhang and Minghui Tang and Meng Li and Miaojun Wang and Mingming Li and Ning Tian and Panpan Huang and Peng Zhang and Qiancheng Wang and Qinyu Chen and Qiushi Du and Ruiqi Ge and Ruisong Zhang and Ruizhe Pan and Runji Wang and R. J. Chen and R. L. Jin and Ruyi Chen and Shanghao Lu and Shangyan Zhou and Shanhuang Chen and Shengfeng Ye and Shiyu Wang and Shuiping Yu and Shunfeng Zhou and Shuting Pan and S. S. Li and Shuang Zhou and Shaoqing Wu and Shengfeng Ye and Tao Yun and Tian Pei and Tianyu Sun and T. Wang and Wangding Zeng and Wanjia Zhao and Wen Liu and Wenfeng Liang and Wenjun Gao and Wenqin Yu and Wentao Zhang and W. L. Xiao and Wei An and Xiaodong Liu and Xiaohan Wang and Xiaokang Chen and Xiaotao Nie and Xin Cheng and Xin Liu and Xin Xie and Xingchao Liu and Xinyu Yang and Xinyuan Li and Xuecheng Su and Xuheng Lin and X. Q. Li and Xiangyue Jin and Xiaojin Shen and Xiaosha Chen and Xiaowen Sun and Xiaoxiang Wang and Xinnan Song and Xinyi Zhou and Xianzu Wang and Xinxia Shan and Y. K. Li and Y. Q. Wang and Y. X. Wei and Yang Zhang and Yanhong Xu and Yao Li and Yao Zhao and Yaofeng Sun and Yaohui Wang and Yi Yu and Yichao Zhang and Yifan Shi and Yiliang Xiong and Ying He and Yishi Piao and Yisong Wang and Yixuan Tan and Yiyang Ma and Yiyuan Liu and Yongqiang Guo and Yuan Ou and Yuduan Wang and Yue Gong and Yuheng Zou and Yujia He and Yunfan Xiong and Yuxiang Luo and Yuxiang You and Yuxuan Liu and Yuyang Zhou and Y. X. Zhu and Yanhong Xu and Yanping Huang and Yaohui Li and Yi Zheng and Yuchen Zhu and Yunxian Ma and Ying Tang and Yukun Zha and Yuting Yan and Z. Z. Ren and Zehui Ren and Zhangli Sha and Zhe Fu and Zhean Xu and Zhenda Xie and Zhengyan Zhang and Zhewen Hao and Zhicheng Ma and Zhigang Yan and Zhiyu Wu and Zihui Gu and Zijia Zhu and Zijun Liu and Zilin Li and Ziwei Xie and Ziyang Song and Zizheng Pan and Zhen Huang and Zhipeng Xu and Zhongyu Zhang and Zhen Zhang},
      year={2025},
      eprint={2501.12948},
      archivePrefix={arXiv},
      primaryClass={cs.CL},
      url={https://arxiv.org/abs/2501.12948}, 
}

@misc{yang2025qwen3technicalreport,
      title={Qwen3 Technical Report}, 
      author={An Yang and Anfeng Li and Baosong Yang and Beichen Zhang and Binyuan Hui and Bo Zheng and Bowen Yu and Chang Gao and Chengen Huang and Chenxu Lv and Chujie Zheng and Dayiheng Liu and Fan Zhou and Fei Huang and Feng Hu and Hao Ge and Haoran Wei and Huan Lin and Jialong Tang and Jian Yang and Jianhong Tu and Jianwei Zhang and Jianxin Yang and Jiaxi Yang and Jing Zhou and Jingren Zhou and Junyang Lin and Kai Dang and Keqin Bao and Kexin Yang and Le Yu and Lianghao Deng and Mei Li and Mingfeng Xue and Mingze Li and Pei Zhang and Peng Wang and Qin Zhu and Rui Men and Ruize Gao and Shixuan Liu and Shuang Luo and Tianhao Li and Tianyi Tang and Wenbiao Yin and Xingzhang Ren and Xinyu Wang and Xinyu Zhang and Xuancheng Ren and Yang Fan and Yang Su and Yichang Zhang and Yinger Zhang and Yu Wan and Yuqiong Liu and Zekun Wang and Zeyu Cui and Zhenru Zhang and Zhipeng Zhou and Zihan Qiu},
      year={2025},
      eprint={2505.09388},
      archivePrefix={arXiv},
      primaryClass={cs.CL},
      url={https://arxiv.org/abs/2505.09388}, 
}

@misc{openai2025gptoss120bgptoss20bmodel,
      title={gpt-oss-120b \& gpt-oss-20b Model Card}, 
      author={OpenAI and : and Sandhini Agarwal and Lama Ahmad and Jason Ai and Sam Altman and Andy Applebaum and Edwin Arbus and Rahul K. Arora and Yu Bai and Bowen Baker and Haiming Bao and Boaz Barak and Ally Bennett and Tyler Bertao and Nivedita Brett and Eugene Brevdo and Greg Brockman and Sebastien Bubeck and Che Chang and Kai Chen and Mark Chen and Enoch Cheung and Aidan Clark and Dan Cook and Marat Dukhan and Casey Dvorak and Kevin Fives and Vlad Fomenko and Timur Garipov and Kristian Georgiev and Mia Glaese and Tarun Gogineni and Adam Goucher and Lukas Gross and Katia Gil Guzman and John Hallman and Jackie Hehir and Johannes Heidecke and Alec Helyar and Haitang Hu and Romain Huet and Jacob Huh and Saachi Jain and Zach Johnson and Chris Koch and Irina Kofman and Dominik Kundel and Jason Kwon and Volodymyr Kyrylov and Elaine Ya Le and Guillaume Leclerc and James Park Lennon and Scott Lessans and Mario Lezcano-Casado and Yuanzhi Li and Zhuohan Li and Ji Lin and Jordan Liss and Lily and Liu and Jiancheng Liu and Kevin Lu and Chris Lu and Zoran Martinovic and Lindsay McCallum and Josh McGrath and Scott McKinney and Aidan McLaughlin and Song Mei and Steve Mostovoy and Tong Mu and Gideon Myles and Alexander Neitz and Alex Nichol and Jakub Pachocki and Alex Paino and Dana Palmie and Ashley Pantuliano and Giambattista Parascandolo and Jongsoo Park and Leher Pathak and Carolina Paz and Ludovic Peran and Dmitry Pimenov and Michelle Pokrass and Elizabeth Proehl and Huida Qiu and Gaby Raila and Filippo Raso and Hongyu Ren and Kimmy Richardson and David Robinson and Bob Rotsted and Hadi Salman and Suvansh Sanjeev and Max Schwarzer and D. Sculley and Harshit Sikchi and Kendal Simon and Karan Singhal and Yang Song and Dane Stuckey and Zhiqing Sun and Philippe Tillet and Sam Toizer and Foivos Tsimpourlas and Nikhil Vyas and Eric Wallace and Xin Wang and Miles Wang and Olivia Watkins and Kevin Weil and Amy Wendling and Kevin Whinnery and Cedric Whitney and Hannah Wong and Lin Yang and Yu Yang and Michihiro Yasunaga and Kristen Ying and Wojciech Zaremba and Wenting Zhan and Cyril Zhang and Brian Zhang and Eddie Zhang and Shengjia Zhao},
      year={2025},
      eprint={2508.10925},
      archivePrefix={arXiv},
      primaryClass={cs.CL},
      url={https://arxiv.org/abs/2508.10925}, 
}

@ARTICLE{wei2022cot,
       author = {{Wei}, Jason and {Wang}, Xuezhi and {Schuurmans}, Dale and {Bosma}, Maarten and {Ichter}, Brian and {Xia}, Fei and {Chi}, Ed and {Le}, Quoc and {Zhou}, Denny},
        title = "{Chain-of-Thought Prompting Elicits Reasoning in Large Language Models}",
      journal = {arXiv e-prints},
         year = 2022,
        month = jan,
          eid = {arXiv:2201.11903},
        pages = {arXiv:2201.11903},
          doi = {10.48550/arXiv.2201.11903},
archivePrefix = {arXiv},
       eprint = {2201.11903},
 primaryClass = {cs.CL},
       adsurl = {https://ui.adsabs.harvard.edu/abs/2022arXiv220111903W}
}

@ARTICLE{openai2024o1,
       author = {{OpenAI} and {:} and {Jaech}, Aaron and {Kalai}, Adam and {Lerer}, Adam and {Richardson}, Adam and {El-Kishky}, Ahmed and {Low}, Aiden and {Helyar}, Alec and {Madry}, Aleksander and {Beutel}, Alex and {Carney}, Alex and {Iftimie}, Alex and {Karpenko}, Alex and {Tachard Passos}, Alex and {Neitz}, Alexander and {Prokofiev}, Alexander and {Wei}, Alexander and {Tam}, Allison and {Bennett}, Ally and {Kumar}, Ananya and {Saraiva}, Andre and {Vallone}, Andrea and {Duberstein}, Andrew and {Kondrich}, Andrew and {Mishchenko}, Andrey and {Applebaum}, Andy and {Jiang}, Angela and {Nair}, Ashvin and {Zoph}, Barret and {Ghorbani}, Behrooz and {Rossen}, Ben and {Sokolowsky}, Benjamin and {Barak}, Boaz and {McGrew}, Bob and {Minaiev}, Borys and {Hao}, Botao and {Baker}, Bowen and {Houghton}, Brandon and {McKinzie}, Brandon and {Eastman}, Brydon and {Lugaresi}, Camillo and {Bassin}, Cary and {Hudson}, Cary and {Li}, Chak Ming and {de Bourcy}, Charles and {Voss}, Chelsea and {Shen}, Chen and {Zhang}, Chong and {Koch}, Chris and {Orsinger}, Chris and {Hesse}, Christopher and {Fischer}, Claudia and {Chan}, Clive and {Roberts}, Dan and {Kappler}, Daniel and {Levy}, Daniel and {Selsam}, Daniel and {Dohan}, David and {Farhi}, David and {Mely}, David and {Robinson}, David and {Tsipras}, Dimitris and {Li}, Doug and {Oprica}, Dragos and {Freeman}, Eben and {Zhang}, Eddie and {Wong}, Edmund and {Proehl}, Elizabeth and {Cheung}, Enoch and {Mitchell}, Eric and {Wallace}, Eric and {Ritter}, Erik and {Mays}, Evan and {Wang}, Fan and {Petroski Such}, Felipe and {Raso}, Filippo and {Leoni}, Florencia and {Tsimpourlas}, Foivos and {Song}, Francis and {von Lohmann}, Fred and {Sulit}, Freddie and {Salmon}, Geoff and {Parascandolo}, Giambattista and {Chabot}, Gildas and {Zhao}, Grace and {Brockman}, Greg and {Leclerc}, Guillaume and {Salman}, Hadi and {Bao}, Haiming and {Sheng}, Hao and {Andrin}, Hart and {Bagherinezhad}, Hessam and {Ren}, Hongyu and {Lightman}, Hunter and {Chung}, Hyung Won and {Kivlichan}, Ian and {O'Connell}, Ian and {Osband}, Ian and {Clavera Gilaberte}, Ignasi and {Akkaya}, Ilge and {Kostrikov}, Ilya and {Sutskever}, Ilya and {Kofman}, Irina and {Pachocki}, Jakub and {Lennon}, James and {Wei}, Jason and {Harb}, Jean and {Twore}, Jerry and {Feng}, Jiacheng and {Yu}, Jiahui and {Weng}, Jiayi and {Tang}, Jie and {Yu}, Jieqi and {Qui{\~n}onero Candela}, Joaquin and {Palermo}, Joe and {Parish}, Joel and {Heidecke}, Johannes and {Hallman}, John and {Rizzo}, John and {Gordon}, Jonathan and {Uesato}, Jonathan and {Ward}, Jonathan and {Huizinga}, Joost and {Wang}, Julie and {Chen}, Kai and {Xiao}, Kai and {Singhal}, Karan and {Nguyen}, Karina and {Cobbe}, Karl and {Shi}, Katy and {Wood}, Kayla and {Rimbach}, Kendra and {Gu-Lemberg}, Keren and {Liu}, Kevin and {Lu}, Kevin and {Stone}, Kevin and {Yu}, Kevin and {Ahmad}, Lama and {Yang}, Lauren and {Liu}, Leo and {Maksin}, Leon and {Ho}, Leyton and {Fedus}, Liam and {Weng}, Lilian and {Li}, Linden and {McCallum}, Lindsay and {Held}, Lindsey and {Kuhn}, Lorenz and {Kondraciuk}, Lukas and {Kaiser}, Lukasz and {Metz}, Luke and {Boyd}, Madelaine and {Trebacz}, Maja and {Joglekar}, Manas and {Chen}, Mark and {Tintor}, Marko and {Meyer}, Mason and {Jones}, Matt and {Kaufer}, Matt and {Schwarzer}, Max and {Shah}, Meghan and {Yatbaz}, Mehmet and {Guan}, Melody Y. and {Xu}, Mengyuan and {Yan}, Mengyuan and {Glaese}, Mia and {Chen}, Mianna and {Lampe}, Michael and {Malek}, Michael and {Wang}, Michele and {Fradin}, Michelle and {McClay}, Mike and {Pavlov}, Mikhail and {Wang}, Miles and {Wang}, Mingxuan and {Murati}, Mira and {Bavarian}, Mo and {Rohaninejad}, Mostafa and {McAleese}, Nat and {Chowdhury}, Neil and {Chowdhury}, Neil and {Ryder}, Nick and {Tezak}, Nikolas and {Brown}, Noam and {Nachum}, Ofir and {Boiko}, Oleg and {Murk}, Oleg and {Watkins}, Olivia and {Chao}, Patrick and {Ashbourne}, Paul and {Izmailov}, Pavel and {Zhokhov}, Peter and {Dias}, Rachel and {Arora}, Rahul and {Lin}, Randall and {Gontijo Lopes}, Rapha and {Gaon}, Raz and {Miyara}, Reah and {Leike}, Reimar},
        title = "{OpenAI o1 System Card}",
      journal = {arXiv e-prints},
         year = 2024,
        month = dec,
          eid = {arXiv:2412.16720},
        pages = {arXiv:2412.16720},
          doi = {10.48550/arXiv.2412.16720},
archivePrefix = {arXiv},
       eprint = {2412.16720},
 primaryClass = {cs.AI},
       adsurl = {https://ui.adsabs.harvard.edu/abs/2024arXiv241216720O}
}

@ARTICLE{yuan2023rft,
       author = {{Yuan}, Zheng and {Yuan}, Hongyi and {Li}, Chengpeng and {Dong}, Guanting and {Lu}, Keming and {Tan}, Chuanqi and {Zhou}, Chang and {Zhou}, Jingren},
        title = "{Scaling Relationship on Learning Mathematical Reasoning with Large Language Models}",
      journal = {arXiv e-prints},
         year = 2023,
        month = aug,
          eid = {arXiv:2308.01825},
        pages = {arXiv:2308.01825},
          doi = {10.48550/arXiv.2308.01825},
archivePrefix = {arXiv},
       eprint = {2308.01825},
 primaryClass = {cs.CL},
       adsurl = {https://ui.adsabs.harvard.edu/abs/2023arXiv230801825Y}
}

@ARTICLE{lambert2024tulu,
       author = {{Lambert}, Nathan and {Morrison}, Jacob and {Pyatkin}, Valentina and {Huang}, Shengyi and {Ivison}, Hamish and {Brahman}, Faeze and {Miranda}, Lester James V. and {Liu}, Alisa and {Dziri}, Nouha and {Lyu}, Shane and {Gu}, Yuling and {Malik}, Saumya and {Graf}, Victoria and {Hwang}, Jena D. and {Yang}, Jiangjiang and {Le Bras}, Ronan and {Tafjord}, Oyvind and {Wilhelm}, Chris and {Soldaini}, Luca and {Smith}, Noah A. and {Wang}, Yizhong and {Dasigi}, Pradeep and {Hajishirzi}, Hannaneh},
        title = "{Tulu 3: Pushing Frontiers in Open Language Model Post-Training}",
      journal = {arXiv e-prints},
         year = 2024,
        month = nov,
          eid = {arXiv:2411.15124},
        pages = {arXiv:2411.15124},
          doi = {10.48550/arXiv.2411.15124},
archivePrefix = {arXiv},
       eprint = {2411.15124},
 primaryClass = {cs.CL},
       adsurl = {https://ui.adsabs.harvard.edu/abs/2024arXiv241115124L}
}

@ARTICLE{Cobbe2021gsm8k,
       author = {{Cobbe}, Karl and {Kosaraju}, Vineet and {Bavarian}, Mohammad and {Chen}, Mark and {Jun}, Heewoo and {Kaiser}, Lukasz and {Plappert}, Matthias and {Tworek}, Jerry and {Hilton}, Jacob and {Nakano}, Reiichiro and {Hesse}, Christopher and {Schulman}, John},
        title = "{Training Verifiers to Solve Math Word Problems}",
      journal = {arXiv e-prints},
         year = 2021,
        month = oct,
          eid = {arXiv:2110.14168},
        pages = {arXiv:2110.14168},
          doi = {10.48550/arXiv.2110.14168},
archivePrefix = {arXiv},
       eprint = {2110.14168},
 primaryClass = {cs.LG},
       adsurl = {https://ui.adsabs.harvard.edu/abs/2021arXiv211014168C}
}

@ARTICLE{ji2025amthinking,
       author = {{Ji}, Yunjie and {Tian}, Xiaoyu and {Zhao}, Sitong and {Wang}, Haotian and {Chen}, Shuaiting and {Peng}, Yiping and {Zhao}, Han and {Li}, Xiangang},
        title = "{AM-Thinking-v1: Advancing the Frontier of Reasoning at 32B Scale}",
      journal = {arXiv e-prints},
         year = 2025,
        month = may,
          eid = {arXiv:2505.08311},
        pages = {arXiv:2505.08311},
          doi = {10.48550/arXiv.2505.08311},
archivePrefix = {arXiv},
       eprint = {2505.08311},
 primaryClass = {cs.CL},
       adsurl = {https://ui.adsabs.harvard.edu/abs/2025arXiv250508311J}
}

@ARTICLE{lightman2023prm800k,
       author = {{Lightman}, Hunter and {Kosaraju}, Vineet and {Burda}, Yura and {Edwards}, Harri and {Baker}, Bowen and {Lee}, Teddy and {Leike}, Jan and {Schulman}, John and {Sutskever}, Ilya and {Cobbe}, Karl},
        title = "{Let's Verify Step by Step}",
      journal = {arXiv e-prints},
         year = 2023,
        month = may,
          eid = {arXiv:2305.20050},
        pages = {arXiv:2305.20050},
          doi = {10.48550/arXiv.2305.20050},
archivePrefix = {arXiv},
       eprint = {2305.20050},
 primaryClass = {cs.LG},
       adsurl = {https://ui.adsabs.harvard.edu/abs/2023arXiv230520050L}
}

@ARTICLE{wang2023mathshepherd,
       author = {{Wang}, Peiyi and {Li}, Lei and {Shao}, Zhihong and {Xu}, R.~X. and {Dai}, Damai and {Li}, Yifei and {Chen}, Deli and {Wu}, Y. and {Sui}, Zhifang},
        title = "{Math-Shepherd: Verify and Reinforce LLMs Step-by-step without Human Annotations}",
      journal = {arXiv e-prints},
         year = 2023,
        month = dec,
          eid = {arXiv:2312.08935},
        pages = {arXiv:2312.08935},
          doi = {10.48550/arXiv.2312.08935},
archivePrefix = {arXiv},
       eprint = {2312.08935},
 primaryClass = {cs.AI},
       adsurl = {https://ui.adsabs.harvard.edu/abs/2023arXiv231208935W}
}

@article{luo2024omegaprm,
       author = {{Luo}, Liangchen and {Liu}, Yinxiao and {Liu}, Rosanne and {Phatale}, Samrat and {Guo}, Meiqi and {Lara}, Harsh and {Li}, Yunxuan and {Shu}, Lei and {Zhu}, Yun and {Meng}, Lei and {Sun}, Jiao and {Rastogi}, Abhinav},
        title = "{Improve Mathematical Reasoning in Language Models by Automated Process Supervision}",
      journal = {arXiv e-prints},
         year = 2024,
        month = jun,
          eid = {arXiv:2406.06592},
        pages = {arXiv:2406.06592},
          doi = {10.48550/arXiv.2406.06592},
archivePrefix = {arXiv},
       eprint = {2406.06592},
 primaryClass = {cs.CL},
       adsurl = {https://ui.adsabs.harvard.edu/abs/2024arXiv240606592L}
}

@article{she2025rprm,
       author = {{She}, Shuaijie and {Liu}, Junxiao and {Liu}, Yifeng and {Chen}, Jiajun and {Huang}, Xin and {Huang}, Shujian},
        title = "{R-PRM: Reasoning-Driven Process Reward Modeling}",
      journal = {arXiv e-prints},
         year = 2025,
        month = mar,
          eid = {arXiv:2503.21295},
        pages = {arXiv:2503.21295},
          doi = {10.48550/arXiv.2503.21295},
archivePrefix = {arXiv},
       eprint = {2503.21295},
 primaryClass = {cs.CL},
       adsurl = {https://ui.adsabs.harvard.edu/abs/2025arXiv250321295S}
}

@ARTICLE{zhang2024genver,
       author = {{Zhang}, Lunjun and {Hosseini}, Arian and {Bansal}, Hritik and {Kazemi}, Mehran and {Kumar}, Aviral and {Agarwal}, Rishabh},
        title = "{Generative Verifiers: Reward Modeling as Next-Token Prediction}",
      journal = {arXiv e-prints},
         year = 2024,
        month = aug,
          eid = {arXiv:2408.15240},
        pages = {arXiv:2408.15240},
          doi = {10.48550/arXiv.2408.15240},
archivePrefix = {arXiv},
       eprint = {2408.15240},
 primaryClass = {cs.LG},
       adsurl = {https://ui.adsabs.harvard.edu/abs/2024arXiv240815240Z}
}

@ARTICLE{mahan2024genrm,
       author = {{Mahan}, Dakota and {Van Phung}, Duy and {Rafailov}, Rafael and {Blagden}, Chase and {Lile}, Nathan and {Castricato}, Louis and {Fr{\"a}nken}, Jan-Philipp and {Finn}, Chelsea and {Albalak}, Alon},
        title = "{Generative Reward Models}",
      journal = {arXiv e-prints},
         year = 2024,
        month = oct,
          eid = {arXiv:2410.12832},
        pages = {arXiv:2410.12832},
          doi = {10.48550/arXiv.2410.12832},
archivePrefix = {arXiv},
       eprint = {2410.12832},
 primaryClass = {cs.LG},
       adsurl = {https://ui.adsabs.harvard.edu/abs/2024arXiv241012832M}
}

@ARTICLE{shi2025heimdall,
       author = {{Shi}, Wenlei and {Jin}, Xing},
        title = "{Heimdall: test-time scaling on the generative verification}",
      journal = {arXiv e-prints},
         year = 2025,
        month = apr,
          eid = {arXiv:2504.10337},
        pages = {arXiv:2504.10337},
          doi = {10.48550/arXiv.2504.10337},
archivePrefix = {arXiv},
       eprint = {2504.10337},
 primaryClass = {cs.AI},
       adsurl = {https://ui.adsabs.harvard.edu/abs/2025arXiv250410337S}
}

@ARTICLE{zheng2024processbench,
       author = {{Zheng}, Chujie and {Zhang}, Zhenru and {Zhang}, Beichen and {Lin}, Runji and {Lu}, Keming and {Yu}, Bowen and {Liu}, Dayiheng and {Zhou}, Jingren and {Lin}, Junyang},
        title = "{ProcessBench: Identifying Process Errors in Mathematical Reasoning}",
      journal = {arXiv e-prints},
         year = 2024,
        month = dec,
          eid = {arXiv:2412.06559},
        pages = {arXiv:2412.06559},
          doi = {10.48550/arXiv.2412.06559},
archivePrefix = {arXiv},
       eprint = {2412.06559},
 primaryClass = {cs.AI},
       adsurl = {https://ui.adsabs.harvard.edu/abs/2024arXiv241206559Z}
}

@misc{yu2025dapoopensourcellmreinforcement,
      title={DAPO: An Open-Source LLM Reinforcement Learning System at Scale}, 
      author={Qiying Yu and Zheng Zhang and Ruofei Zhu and Yufeng Yuan and Xiaochen Zuo and Yu Yue and Weinan Dai and Tiantian Fan and Gaohong Liu and Lingjun Liu and Xin Liu and Haibin Lin and Zhiqi Lin and Bole Ma and Guangming Sheng and Yuxuan Tong and Chi Zhang and Mofan Zhang and Wang Zhang and Hang Zhu and Jinhua Zhu and Jiaze Chen and Jiangjie Chen and Chengyi Wang and Hongli Yu and Yuxuan Song and Xiangpeng Wei and Hao Zhou and Jingjing Liu and Wei-Ying Ma and Ya-Qin Zhang and Lin Yan and Mu Qiao and Yonghui Wu and Mingxuan Wang},
      year={2025},
      eprint={2503.14476},
      archivePrefix={arXiv},
      primaryClass={cs.LG},
      url={https://arxiv.org/abs/2503.14476}, 
}

@ARTICLE{lyu2025oreal,
       author = {{Lyu}, Chengqi and {Gao}, Songyang and {Gu}, Yuzhe and {Zhang}, Wenwei and {Gao}, Jianfei and {Liu}, Kuikun and {Wang}, Ziyi and {Li}, Shuaibin and {Zhao}, Qian and {Huang}, Haian and {Cao}, Weihan and {Liu}, Jiangning and {Liu}, Hongwei and {Liu}, Junnan and {Zhang}, Songyang and {Lin}, Dahua and {Chen}, Kai},
        title = "{Exploring the Limit of Outcome Reward for Learning Mathematical Reasoning}",
      journal = {arXiv e-prints},
         year = 2025,
        month = feb,
          eid = {arXiv:2502.06781},
        pages = {arXiv:2502.06781},
          doi = {10.48550/arXiv.2502.06781},
archivePrefix = {arXiv},
       eprint = {2502.06781},
 primaryClass = {cs.CL},
       adsurl = {https://ui.adsabs.harvard.edu/abs/2025arXiv250206781L}
}

@ARTICLE{liu2025compassverifier,
       author = {{Liu}, Shudong and {Liu}, Hongwei and {Liu}, Junnan and {Xiao}, Linchen and {Gao}, Songyang and {Lyu}, Chengqi and {Gu}, Yuzhe and {Zhang}, Wenwei and {Wong}, Derek F. and {Zhang}, Songyang and {Chen}, Kai},
        title = "{CompassVerifier: A Unified and Robust Verifier for LLMs Evaluation and Outcome Reward}",
      journal = {arXiv e-prints},
         year = 2025,
        month = aug,
          eid = {arXiv:2508.03686},
        pages = {arXiv:2508.03686},
          doi = {10.48550/arXiv.2508.03686},
archivePrefix = {arXiv},
       eprint = {2508.03686},
 primaryClass = {cs.CL},
       adsurl = {https://ui.adsabs.harvard.edu/abs/2025arXiv250803686L}
}

@article{zhang2025tts,
       author = {{Zhang}, Qiyuan and {Lyu}, Fuyuan and {Sun}, Zexu and {Wang}, Lei and {Zhang}, Weixu and {Hua}, Wenyue and {Wu}, Haolun and {Guo}, Zhihan and {Wang}, Yufei and {Muennighoff}, Niklas and {King}, Irwin and {Liu}, Xue and {Ma}, Chen},
        title = "{A Survey on Test-Time Scaling in Large Language Models: What, How, Where, and How Well?}",
      journal = {arXiv e-prints},
         year = 2025,
        month = mar,
          eid = {arXiv:2503.24235},
        pages = {arXiv:2503.24235},
          doi = {10.48550/arXiv.2503.24235},
archivePrefix = {arXiv},
       eprint = {2503.24235},
 primaryClass = {cs.CL},
       adsurl = {https://ui.adsabs.harvard.edu/abs/2025arXiv250324235Z}
}

@article{he2024olympiadbench,
       author = {{He}, Chaoqun and {Luo}, Renjie and {Bai}, Yuzhuo and {Hu}, Shengding and {Leng Thai}, Zhen and {Shen}, Junhao and {Hu}, Jinyi and {Han}, Xu and {Huang}, Yujie and {Zhang}, Yuxiang and {Liu}, Jie and {Qi}, Lei and {Liu}, Zhiyuan and {Sun}, Maosong},
        title = "{OlympiadBench: A Challenging Benchmark for Promoting AGI with Olympiad-Level Bilingual Multimodal Scientific Problems}",
      journal = {arXiv e-prints},
         year = 2024,
        month = feb,
          eid = {arXiv:2402.14008},
        pages = {arXiv:2402.14008},
          doi = {10.48550/arXiv.2402.14008},
archivePrefix = {arXiv},
       eprint = {2402.14008},
 primaryClass = {cs.CL},
       adsurl = {https://ui.adsabs.harvard.edu/abs/2024arXiv240214008H}
}

@ARTICLE{hendryck2021math,
       author = {{Hendrycks}, Dan and {Burns}, Collin and {Kadavath}, Saurav and {Arora}, Akul and {Basart}, Steven and {Tang}, Eric and {Song}, Dawn and {Steinhardt}, Jacob},
        title = "{Measuring Mathematical Problem Solving With the MATH Dataset}",
      journal = {arXiv e-prints},
         year = 2021,
        month = mar,
          eid = {arXiv:2103.03874},
        pages = {arXiv:2103.03874},
          doi = {10.48550/arXiv.2103.03874},
archivePrefix = {arXiv},
       eprint = {2103.03874},
 primaryClass = {cs.LG},
       adsurl = {https://ui.adsabs.harvard.edu/abs/2021arXiv210303874H}
}

@article{zhang2025lessons,
       author = {{Zhang}, Zhenru and {Zheng}, Chujie and {Wu}, Yangzhen and {Zhang}, Beichen and {Lin}, Runji and {Yu}, Bowen and {Liu}, Dayiheng and {Zhou}, Jingren and {Lin}, Junyang},
        title = "{The Lessons of Developing Process Reward Models in Mathematical Reasoning}",
      journal = {arXiv e-prints},
         year = 2025,
        month = jan,
          eid = {arXiv:2501.07301},
        pages = {arXiv:2501.07301},
          doi = {10.48550/arXiv.2501.07301},
archivePrefix = {arXiv},
       eprint = {2501.07301},
 primaryClass = {cs.CL},
       adsurl = {https://ui.adsabs.harvard.edu/abs/2025arXiv250107301Z}
}

@article{yang2025beyondthefirsterror,
       author = {{Yang}, Zhaohui and {He}, Chenghua and {Shi}, Xiaowen and {Li}, Linjing and {Yin}, Qiyue and {Deng}, Shihong and {Jiang}, Daxin},
        title = "{Beyond the First Error: Process Reward Models for Reflective Mathematical Reasoning}",
      journal = {arXiv e-prints},
         year = 2025,
        month = may,
          eid = {arXiv:2505.14391},
        pages = {arXiv:2505.14391},
          doi = {10.48550/arXiv.2505.14391},
archivePrefix = {arXiv},
       eprint = {2505.14391},
 primaryClass = {cs.AI},
       adsurl = {https://ui.adsabs.harvard.edu/abs/2025arXiv250514391Y}
}

@article{duan2025actprm,
       author = {{Duan}, Keyu and {Liu}, Zichen and {Mao}, Xin and {Pang}, Tianyu and {Chen}, Changyu and {Chen}, Qiguang and {Qizhe Shieh}, Michael and {Dou}, Longxu},
        title = "{Efficient Process Reward Model Training via Active Learning}",
      journal = {arXiv e-prints},
         year = 2025,
        month = apr,
          eid = {arXiv:2504.10559},
        pages = {arXiv:2504.10559},
          doi = {10.48550/arXiv.2504.10559},
archivePrefix = {arXiv},
       eprint = {2504.10559},
 primaryClass = {stat.ML},
       adsurl = {https://ui.adsabs.harvard.edu/abs/2025arXiv250410559D}
}

@misc{anthony2017thinkingfastslowdeep,
      title={Thinking Fast and Slow with Deep Learning and Tree Search}, 
      author={Thomas Anthony and Zheng Tian and David Barber},
      year={2017},
      eprint={1705.08439},
      archivePrefix={arXiv},
      primaryClass={cs.AI},
      url={https://arxiv.org/abs/1705.08439}, 
}

@ARTICLE{deepseekv3,
       author = {{DeepSeek-AI} and {Liu}, Aixin and {Feng}, Bei and {Xue}, Bing and {Wang}, Bingxuan and {Wu}, Bochao and {Lu}, Chengda and {Zhao}, Chenggang and {Deng}, Chengqi and {Zhang}, Chenyu and {Ruan}, Chong and {Dai}, Damai and {Guo}, Daya and {Yang}, Dejian and {Chen}, Deli and {Ji}, Dongjie and {Li}, Erhang and {Lin}, Fangyun and {Dai}, Fucong and {Luo}, Fuli and {Hao}, Guangbo and {Chen}, Guanting and {Li}, Guowei and {Zhang}, H. and {Bao}, Han and {Xu}, Hanwei and {Wang}, Haocheng and {Zhang}, Haowei and {Ding}, Honghui and {Xin}, Huajian and {Gao}, Huazuo and {Li}, Hui and {Qu}, Hui and {Cai}, J.~L. and {Liang}, Jian and {Guo}, Jianzhong and {Ni}, Jiaqi and {Li}, Jiashi and {Wang}, Jiawei and {Chen}, Jin and {Chen}, Jingchang and {Yuan}, Jingyang and {Qiu}, Junjie and {Li}, Junlong and {Song}, Junxiao and {Dong}, Kai and {Hu}, Kai and {Gao}, Kaige and {Guan}, Kang and {Huang}, Kexin and {Yu}, Kuai and {Wang}, Lean and {Zhang}, Lecong and {Xu}, Lei and {Xia}, Leyi and {Zhao}, Liang and {Wang}, Litong and {Zhang}, Liyue and {Li}, Meng and {Wang}, Miaojun and {Zhang}, Mingchuan and {Zhang}, Minghua and {Tang}, Minghui and {Li}, Mingming and {Tian}, Ning and {Huang}, Panpan and {Wang}, Peiyi and {Zhang}, Peng and {Wang}, Qiancheng and {Zhu}, Qihao and {Chen}, Qinyu and {Du}, Qiushi and {Chen}, R.~J. and {Jin}, R.~L. and {Ge}, Ruiqi and {Zhang}, Ruisong and {Pan}, Ruizhe and {Wang}, Runji and {Xu}, Runxin and {Zhang}, Ruoyu and {Chen}, Ruyi and {Li}, S.~S. and {Lu}, Shanghao and {Zhou}, Shangyan and {Chen}, Shanhuang and {Wu}, Shaoqing and {Ye}, Shengfeng and {Ye}, Shengfeng and {Ma}, Shirong and {Wang}, Shiyu and {Zhou}, Shuang and {Yu}, Shuiping and {Zhou}, Shunfeng and {Pan}, Shuting and {Wang}, T. and {Yun}, Tao and {Pei}, Tian and {Sun}, Tianyu and {Xiao}, W.~L. and {Zeng}, Wangding and {Zhao}, Wanjia and {An}, Wei and {Liu}, Wen and {Liang}, Wenfeng and {Gao}, Wenjun and {Yu}, Wenqin and {Zhang}, Wentao and {Li}, X.~Q. and {Jin}, Xiangyue and {Wang}, Xianzu and {Bi}, Xiao and {Liu}, Xiaodong and {Wang}, Xiaohan and {Shen}, Xiaojin and {Chen}, Xiaokang and {Zhang}, Xiaokang and {Chen}, Xiaosha and {Nie}, Xiaotao and {Sun}, Xiaowen and {Wang}, Xiaoxiang and {Cheng}, Xin and {Liu}, Xin and {Xie}, Xin and {Liu}, Xingchao and {Yu}, Xingkai and {Song}, Xinnan and {Shan}, Xinxia and {Zhou}, Xinyi and {Yang}, Xinyu and {Li}, Xinyuan and {Su}, Xuecheng and {Lin}, Xuheng and {Li}, Y.~K. and {Wang}, Y.~Q. and {Wei}, Y.~X. and {Zhu}, Y.~X. and {Zhang}, Yang and {Xu}, Yanhong and {Xu}, Yanhong and {Huang}, Yanping and {Li}, Yao and {Zhao}, Yao and {Sun}, Yaofeng and {Li}, Yaohui and {Wang}, Yaohui and {Yu}, Yi and {Zheng}, Yi and {Zhang}, Yichao and {Shi}, Yifan and {Xiong}, Yiliang and {He}, Ying and {Tang}, Ying and {Piao}, Yishi and {Wang}, Yisong and {Tan}, Yixuan and {Ma}, Yiyang and {Liu}, Yiyuan and {Guo}, Yongqiang and {Wu}, Yu and {Ou}, Yuan and {Zhu}, Yuchen and {Wang}, Yuduan and {Gong}, Yue and {Zou}, Yuheng and {He}, Yujia and {Zha}, Yukun and {Xiong}, Yunfan and {Ma}, Yunxian and {Yan}, Yuting and {Luo}, Yuxiang and {You}, Yuxiang and {Liu}, Yuxuan and {Zhou}, Yuyang and {Wu}, Z.~F. and {Ren}, Z.~Z. and {Ren}, Zehui and {Sha}, Zhangli and {Fu}, Zhe and {Xu}, Zhean and {Huang}, Zhen and {Zhang}, Zhen and {Xie}, Zhenda and {Zhang}, Zhengyan and {Hao}, Zhewen and {Gou}, Zhibin and {Ma}, Zhicheng and {Yan}, Zhigang and {Shao}, Zhihong and {Xu}, Zhipeng and {Wu}, Zhiyu and {Zhang}, Zhongyu and {Li}, Zhuoshu and {Gu}, Zihui and {Zhu}, Zijia and {Liu}, Zijun and {Li}, Zilin and {Xie}, Ziwei and {Song}, Ziyang and {Gao}, Ziyi and {Pan}, Zizheng},
        title = "{DeepSeek-V3 Technical Report}",
      journal = {arXiv e-prints},
         year = 2024,
        month = dec,
          eid = {arXiv:2412.19437},
        pages = {arXiv:2412.19437},
          doi = {10.48550/arXiv.2412.19437},
archivePrefix = {arXiv},
       eprint = {2412.19437},
 primaryClass = {cs.CL},
       adsurl = {https://ui.adsabs.harvard.edu/abs/2024arXiv241219437D}
}

@article{li2024numinamath,
  title={Numinamath: The largest public dataset in ai4maths with 860k pairs of competition math problems and solutions},
  author={Li, Jia and Beeching, Edward and Tunstall, Lewis and Lipkin, Ben and Soletskyi, Roman and Huang, Shengyi and Rasul, Kashif and Yu, Longhui and Jiang, Albert Q and Shen, Ziju and others},
  journal={Hugging Face repository},
  volume={13},
  number={9},
  pages={9},
  year={2024}
}

@article{mahdavi2025leveraging,
  title={Leveraging online olympiad-level math problems for llms training and contamination-resistant evaluation},
  author={Mahdavi, Sadegh and Li, Muchen and Liu, Kaiwen and Thrampoulidis, Christos and Sigal, Leonid and Liao, Renjie},
  journal={arXiv preprint arXiv:2501.14275},
  year={2025}
}

@article{gao2024omni,
  title={Omni-math: A universal olympiad level mathematic benchmark for large language models},
  author={Gao, Bofei and Song, Feifan and Yang, Zhe and Cai, Zefan and Miao, Yibo and Dong, Qingxiu and Li, Lei and Ma, Chenghao and Chen, Liang and Xu, Runxin and others},
  journal={arXiv preprint arXiv:2410.07985},
  year={2024}
}


\clearpage
\appendix
\newpage

\section*{The Use of Large Language Models (LLMs)}
We used LLMs solely for language polishing. The scientific ideas, methodology, analyses, and conclusions were entirely developed by the authors, while the LLMs assisted only in improving clarity and readability of the text.

\section{Annotation Details}
\label{appendix:annotation details}

\subsection{Data and Annotation Preliminaries}
\label{subsec: annotation protocol}



\textbf{Problem Curation.}
We curated math problems from widely-used benchmarks, published problem sets and open contests to ensure a broad spectrum of challenges in terms of difficulty and knowledge domains. The problems span K-12 education, high-school competitions, and undergraduate-level mathematics. To construct our dataset, we aggregated our initial data pool from three primary sources: (1) \textbf{Challenging Benchmarks ($\sim$1k samples):} High-difficulty problems from Putnam and USAMO; (2) \textbf{Math Forums ($\sim$10k samples):} Community-curated advanced problems from AoPS\footnote{\url{https://artofproblemsolving.com/community}}; and (3) \textbf{Open-Source Datasets ($\sim$1M samples):} Large-scale collections including NuminaMath \citep{li2024numinamath} and AoPS-Instruct \citep{mahdavi2025leveraging}. We explicitly excluded multiple-choice, fill-in-the-blank, and true/false questions, as these formats allow for correct answers via "lucky guessing" or shortcuts. Consequently, we retained only open-ended calculation and proof-based problems that necessitate rigorous, step-by-step reasoning. Furthermore, to prevent data leakage, we implemented a strict decontamination pipeline against major public benchmarks (GSM8K \citep{Cobbe2021gsm8k}, MATH \citep{hendryck2021math}, OlympiadBench \citep{he2024olympiadbench}, Omni-MATH \citep{gao2024omni}, AIME 2024, and AIME 2025). This process utilized a two-stage mechanism: first, Exact Matching via normalized string comparison to remove duplicates; and second, Semantic Matching using LLM-based embedding similarity to identify and discard queries semantically similar to the evaluation sets.

\textbf{CoT Generation and Summarization.}
We sample 8 to 12 unique CoTs per problem from state-of-the-art models (R1 and Qwen families) to capture diverse reasoning paths. Initial attempts to use the default summaries following the \texttt{</think>} tag proved inadequate, as they omitted crucial intermediate steps and resisted improvement through prompt engineering. Therefore, we employ Deepseek-V3 to re-summarize the reasoning content within \texttt{<think>...</think>} tags, preserving all calculations, enumerations, and case analyses while segmenting steps uniformly with "\verb|---|" delimiters. This procedure yields our initial unlabeled data pool, $\mathcal{D_U} = \{ (P_1, \mathcal{S}_1), \ldots, (P_m, \mathcal{S}_m) \}$, which serves as the starting point for active learning. 

\textbf{Active Learning Configuration}
To maximize the utility of the annotation budget, we employed a mixed sampling strategy. In each iteration loop, the data selected for expert annotation consists of two parts: (1) \textbf{Uncertainty Sampling (80\%):} The majority of the budget is allocated to samples with the lowest consistency scores. We utilized a dynamic consistency threshold $\tau_t$, set to $\tau_1=0.25$ for the first stage and $\tau_2=0.5$ for the second stage, to capture increasingly subtle errors as the model improves; (2) \textbf{High-Confidence Sampling (20\%):} To prevent the model from becoming overconfident or forgetting established knowledge, the remaining 20\% of samples are randomly selected from the high-consistency pool. This ensures the verifier receives feedback on cases it considers "correct," allowing us to correct confident errors (false negatives) during the Expert Iteration phase.

\textbf{Annotation Protocol.}
We established a precise protocol to guide expert annotation. For each sample $(P_i, \mathcal{S}_i)$, annotators provide a brief explanation $\hat{\mathcal{E}_i}$ and identify the index $\hat{\ell_i}$ of the first erroneous step. Reference solutions were provided to facilitate this process.
To address ambiguity, we instructed annotators to identify "flawed but tolerable" steps—steps that are imperfect but could be easily corrected within 2-3 sentences and precede the first definitive error. Such steps were not classified as erroneous.
To ensure data quality, three experts independently evaluate each solution. Annotations are valid only when: (1) all experts agree the solution is correct, or (2) at least two experts identify an error within a two-step window. This window accounts for errors that span multiple steps and resist single-step attribution. As shown in Tab.~\ref{tab:stats}, annotators typically achieve stronger consensus.
This protocol ensures only high-confidence labels are added to our dataset during active learning.

\subsection{Dataset Statistics}
\label{appendix sec:dataset statistics}

\begin{figure}[!ht]
  \centering
  \begin{minipage}[b]{0.48\textwidth}
    \centering
    \includegraphics[width=\linewidth]{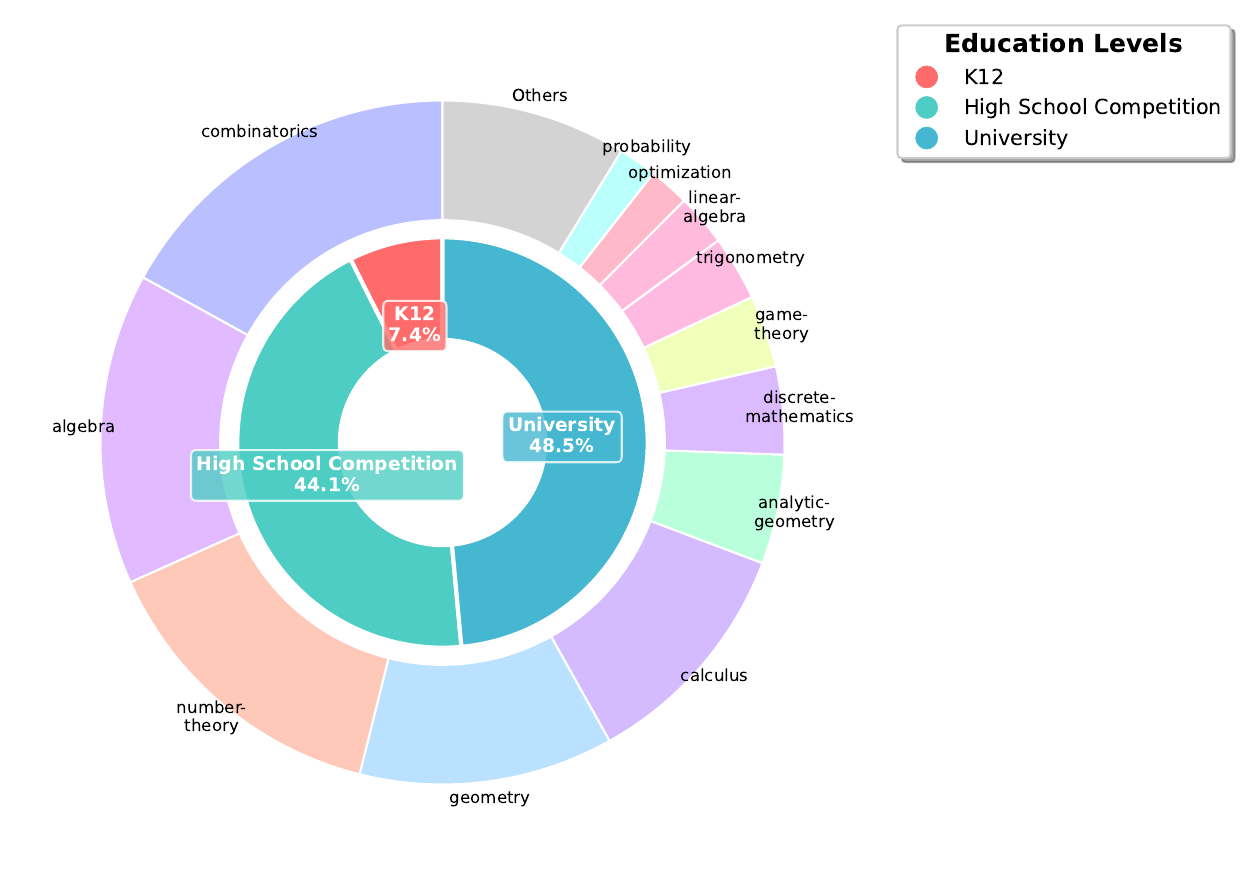}
    \caption{The topic distribution by chart of categories (inner) and domains (outer).}
    \label{fig:topic_distribution}
  \end{minipage}
  \hfill
  \begin{minipage}[b]{0.48\textwidth}
    \centering
    \includegraphics[width=0.8\linewidth]{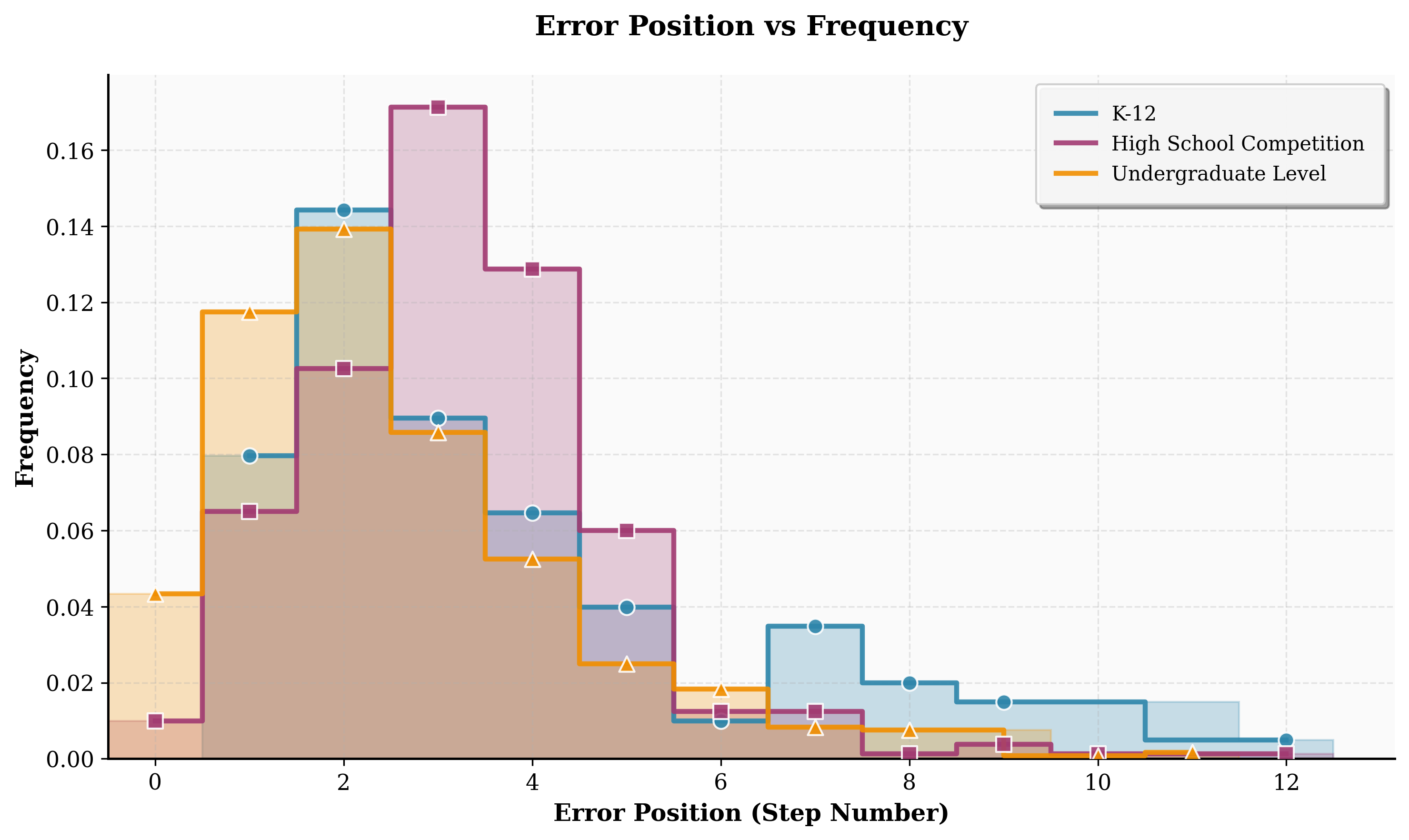}
    \caption{Distribution of error positions in \textsc{\thisbench}.}
    \label{fig:error_position_dist}
  \end{minipage}
\end{figure}

\begin{table}[!t]
  \centering
  \caption{Statistics of \textsc{\thisbench}.
``\% $\ge n$ steps'' denotes the proportion of samples whose solutions have $\ge n$ step.
``\% 3/3 and 2/3 agreement'' denotes the proportion of samples for which each type of annotator agreement is achieved.
  }
  \vspace{-1mm}
  \scalebox{0.95}{
    \begin{tabular}{lcccccc}
    \toprule
     & \multicolumn{2}{c}{\textbf{K12}} & \multicolumn{2}{c}{\textbf{Highschool Competition}} & \multicolumn{2}{c}{\textbf{Undergraduate}} \\
    \cmidrule(lr){2-3}\cmidrule(lr){4-5}\cmidrule(lr){6-7}
    & \textbf{error} & \textbf{correct} & \textbf{error} & \textbf{correct} & \textbf{error} & \textbf{correct} \\
    \midrule
    \multirow{1}[0]{*}{\# Samples} & 82   & 119   & 398   & 402   & 600   & 600  \\
    \midrule

    \# Average Steps          & 6.74   &  6.70  & 7.43   & 7.11   & 6.73   & 6.71  \\

    \% $\ge$ 4 steps & 93.9\% & 97.5\% & 100.0\% & 100.0\% & 99.0\% & 97.3\% \\
    \% $\ge$ 8 steps & 30.5\% & 30.3\% & 40.2\% & 36.1\% & 29.5\% & 29.3\% \\
    \% $\ge$ 12 steps & 6.1\% & 5.0\% & 3.5\% & 2.5\% & 1.3\% & 3.2\% \\
    \midrule
    \# Average Error Position & 3.25   &  /  & 2.97   & /   & 2.49   & /  \\
    \midrule
    \% 3/3 agreement & 76.8\% & 100.0\% & 85.4\% & 100.0\% & 100.0\% & 100.0\% \\
    \% 2/3 agreement & 23.2\% & 0.0\% & 14.6\% & 0.0\% & 0.0\% & 0.0\% \\
    \bottomrule
    \end{tabular}%
  }
  \vspace{2mm}
  \label{tab:stats}%
\end{table}%

The resulting \textsc{\thisbench} comprises three subsets totaling 2,202 test cases. Tab.~\ref{tab:stats} presents detailed statistics, and Fig.~\ref{fig:error_position_dist} shows the error position distribution within erroneous samples. Since all solutions are re-summarized by Deepseek-V3 before evaluation, average step counts remain consistent across problems within each subset. However, incorrect solutions contain slightly more steps than correct ones. Most errors occur in the initial reasoning steps. High school competition problems, which require less formula application but more exploration and analysis, show a delayed error peak compared to other problems.

\section{Verifier Training Details}
\label{appendix sec: implement_detail}

In our experimental framework, we employ the pre-trained R1-distilled-32B~\cite{deepseekai2025deepseekr1incentivizingreasoningcapability} model as the base architecture for fine-tuning the verifier. The verifier is trained with the following hyperparameters: 1 epoch, a learning rate of 8e-5, a sequence length of 32k, and a weight decay of 10. We utilize the DAPO training algorithm with specific configurations for the final on-policy RL stage. For the generation settings of the RL stage, we configure a global batch size of 256 and a prompt repetition factor of 8 to generate multiple samples per prompt. The optimization utilizes a global batch size of 128. Training spans 80 total steps without warmup.

\section{Prompting Details}
\label{appendix: prompt}

The re-summarization prompt and verification prompt are shown in Fig. \ref{fig:prompt_template_solution_reformatting} and Fig.\ref{fig:prompt_template_critique}.

\begin{figure}[!ht] 
\begin{AIbox}{}
As a student proficient in mathematics, you are presented with a problem and a standard solution.

Your task is to transform the given standard solution into a clear, well-structured solution that precisely articulates each reasoning step based on the material provided. Your final solution should be comprehensive, methodical, and leave no logical gaps. It should be in Chinese and should be split into steps divided by the "---" delimiter. Remember to use the same notation, symbols and language as the original solution.

If the standard solution is very brief, you may present it in just 2-3 steps. Otherwise, aim for 5-15 steps.

1. Solution Analysis: Carefully review the provided standard solution to understand the complete solution path.

2. Structured Step-by-Step Solution: Provide a thorough solution where each step follows logically from the previous one. Clearly explain every mathematical operation, theorem application, and reasoning transition mentioned in the standard solution.

3. Strict Adherence to Original Content: Your steps must strictly follow the information provided in the standard solution. Do not add new methods, concepts, or information not found in the original solution. Your task is to reorganize and clarify the existing content, not to enhance or expand it.

4. Mathematical Rigor: When presenting formulas, equations, or mathematical statements, ensure they are expressed with proper notation and justified appropriately. Define all variables and symbols when they first appear.

5. Format Requirements: Present your solution as a sequence of paragraphs, each thoroughly describing a distinct step in the solution process. Separate paragraphs with "---" as delimiters. Do not use numbered or bulleted lists. Each paragraph should represent a meaningful unit of the solution with comparable depth and detail. If the problem contains multiple sub-problems, maintain this paragraph structure throughout, treating the entire solution as a continuous sequence of steps. Conclude with a final paragraph summarizing the answer if appropriate. A sub-problem should have at least 2 paragraphs splitted. Do not use any lists or indexes nested inside a paragraph separated by "---". Since the solution is separated by the "---" delimiters that implicitly defines the index of the steps, you should not introduce any other explicit indexes, neither in numbers, English or Chinese. Every "---" delimiter should occupy a whole line.

Mathematical expressions should be clearly formatted, with proper attention to notation, subscripts, superscripts, and mathematical symbols. Use standard mathematical notation conventions.

PROBLEM STATEMENT:
\{problem\_statement\}

STANDARD SOLUTION:
\{standard\_solution\}

DETAILED SOLUTION WITH STEP-BY-STEP REASONING AND PROPER FORMAT:
\end{AIbox} 
\caption{{\textbf{Prompt Template for Summarization.} }}

\label{fig:prompt_template_solution_reformatting}
\end{figure}

\begin{figure}[!ht] 
\begin{AIbox}{}
You are a mathematics and educational expert tasked with evaluating the correctness of a student's answer. The student's solution is broken down into steps, and your goal is to identify the index of the first incorrect step. The index starts at zero for the first step. If all steps are correct, you should output -1.

Instructions:
- You will receive a question along with the student's answer, divided into steps. Each step is presented in a separate paragraph.

- You are encouraged to express your internal reasoning, but your final response must always include an integer within \textbackslash{}box\{STEP\}. For example, if step 2 is incorrect, respond with \textbackslash{}box\{STEP2\}. If all steps are correct, respond with \textbackslash{}box\{STEP-1\}. Also, after you reported the incorrect step index, you should also briefly report the reason for this incorrectness.

- Some steps may initially appear incorrect but are later corrected in subsequent steps. If a reflection or revision is both accurate and reasonable, the step should be considered correct. If there are multiple reflections, consider only the final one. 

- The student's answer may employ multiple approaches to solve the problem. Within a single response, some approaches may be logically sound while others may not be. If the final conclusions are correct and at least one approach is logically sound, the entire solution should be considered correct.

- In cases where the problem is ambiguous, consider all possible interpretations and determine if the student's response aligns with any of them.

- Evaluate the entire solution, as some intermediate steps might seem incorrect initially but are rectified later, such as dismissing an extraneous root. Ensure you consider the entire context and, if necessary, review the steps more than once.

- The errors to identify can be very subtle, sometimes hiding in the inexplicit applications of theorems or conditions. So you should actively checking every small logical inferences at a small granularity carefully, either in natural language or in formulas. 

- To help you identify the possible errors, every first time you checking a step, you should repeat it in case you missed subtle information. Then you should check its validity by examining its logical inferences within the step/sentences/sub-sentences one by one. When a proof is required, meticulously verify the soundness of each logical step. Incomplete inductive reasoning is unacceptable. If you suspect a step is flawed, consider constructing counter-examples to test its validity.

- Every step should have solid logical basis. Guessing without proof is not allowed.

- For you convenience, you may be provided with a reference solution. The reference solution might be gaped or just a hint. It rarely is wrong but do not miss the possibility. However, some problems might ask just for one valid answer while multiple possible answers exist. In such case you should not judge the students answer is wrong because it has a different solution. Reference answers should be viewed as guides rather than absolute standards. Students may use alternative methods, notations, or approaches that are equally valid.

---

**Question**: \{question\}

**Answer to Verify**: \{answer\_to\_verify\}

\end{AIbox} 
\caption{\modified{\textbf{Prompt Template for Process Verification.} }}

\label{fig:prompt_template_critique}
\end{figure}

\clearpage
\section{Additional Experiment Results}
\label{appendix sec: detailed results}

\begin{table}[H]
  \centering
  \vspace{-1mm}
  \caption{
  Detailed evaluation results on GSM8k partition of ProcessBench.
  }
  \vspace{-1mm}
  \resizebox{0.9\linewidth}{!}{
    \begin{tabular}{lcccccccccccc}
    \toprule
    \multirow{2}{*}{\textbf{Model}} & \multicolumn{4}{c}{\textbf{Absolute}} & \multicolumn{4}{c}{\textbf{Approximate}} & \multicolumn{4}{c}{\textbf{Rough}} \\
    \cmidrule(lr){2-5} \cmidrule(lr){6-9} \cmidrule(lr){10-13}
      & \textbf{Acc} & \textbf{Pre} & \textbf{Rec} & \textbf{F1} & \textbf{Acc} & \textbf{Pre} & \textbf{Rec} & \textbf{F1} & \textbf{Acc} & \textbf{Pre} & \textbf{Rec} & \textbf{F1} \\
    \midrule
    \multicolumn{13}{c}{\textit{\textbf{With Standard Answers}}} \\
    \midrule
    DeepSeek-V3-0324        & 88.5 & 81.0 & 99.5 & 89.3 & 94.5 & 90.1 & 99.5 & 94.6 & 97.8 & 96.0 & 99.5 & 97.7 \\
    DeepSeek-R1-0528        & 92.0 & 86.4 & 99.0 & 92.3 & 96.0 & 93.2 & 99.0 & 96.0 & 98.5 & 97.9 & 99.0 & 98.5 \\
    Qwen-Max-Preview        & 91.3 & 85.0 & 99.5 & 91.6 & 95.0 & 91.0 & 99.5 & 95.0 & 98.5 & 97.5 & 99.5 & 98.5 \\
    gpt-oss-120b(high)      & 91.5 & 85.3 & 99.5 & 91.9 & 94.5 & 90.1 & 99.5 & 94.6 & 98.3 & 97.0 & 99.5 & 98.2 \\
    Qwen2.5-Math-PRM-72B & 89.0 & 82.0 & 99.0 & 89.7 & 94.5 & 90.5 & 99.0 & 94.6 & 97.8 & 96.5 & 99.0 & 97.7 \\
    DeepSeek-R1-Distill-32B & 87.8 & 81.6 & 96.4 & 88.4 & 92.5 & 89.0 & 96.4 & 92.5 & 96.0 & 95.4 & 96.4 & 95.9 \\
    OPV                     & 91.0 & 84.6 & 99.5 & 91.4 & 94.5 & 90.1 & 99.5 & 94.6 & 98.5 & 97.5 & 99.5 & 98.5 \\
    \midrule
    \multicolumn{13}{c}{\textit{\textbf{Without Standard Answers}}} \\
    \midrule
    DeepSeek-V3-0324        & 84.3 & 76.2 & 97.9 & 85.7 & 88.5 & 81.8 & 97.9 & 89.2 & 91.3 & 85.9 & 97.9 & 91.5 \\
    DeepSeek-R1-0528        & 91.5 & 86.6 & 97.4 & 91.7 & 95.0 & 92.6 & 97.4 & 94.9 & 97.3 & 96.9 & 97.4 & 97.2 \\
    Qwen-Max-Preview        & 91.8 & 86.4 & 98.4 & 92.0 & 94.8 & 91.3 & 98.4 & 94.8 & 97.0 & 95.5 & 98.4 & 96.9 \\
    gpt-oss-120b(high)      & 92.3 & 87.5 & 97.9 & 92.4 & 95.8 & 93.6 & 97.9 & 95.7 & 98.0 & 97.9 & 97.9 & 97.9 \\
    Qwen2.5-Math-PRM-72B & 87.3 & 79.8 & 98.5 & 88.2 & 92.0 & 86.8 & 98.5 & 92.2 & 94.5 & 90.9 & 98.5 & 94.5 \\
    DeepSeek-R1-Distill-32B & 89.3 & 82.3 & 99.0 & 89.9 & 92.0 & 86.4 & 99.0 & 92.3 & 94.3 & 90.1 & 99.0 & 94.3 \\
    OPV                     & 89.8 & 83.9 & 97.4 & 90.2 & 94.0 & 90.8 & 97.4 & 94.0 & 97.3 & 96.9 & 97.4 & 97.2 \\
    \bottomrule
    \end{tabular}%
  }
  \label{tab:gsm8k_results}%
\end{table}

\begin{table}[H]
  \centering
  \vspace{-1mm}
  \caption{
  Detailed evaluation results on MATH partition of ProcessBench.
  }
  \vspace{-1mm}
  \resizebox{0.9\linewidth}{!}{
    \begin{tabular}{lcccccccccccc}
    \toprule
    \multirow{2}{*}{\textbf{Model}} & \multicolumn{4}{c}{\textbf{Absolute}} & \multicolumn{4}{c}{\textbf{Approximate}} & \multicolumn{4}{c}{\textbf{Rough}} \\
    \cmidrule(lr){2-5} \cmidrule(lr){6-9} \cmidrule(lr){10-13}
      & \textbf{Acc} & \textbf{Pre} & \textbf{Rec} & \textbf{F1} & \textbf{Acc} & \textbf{Pre} & \textbf{Rec} & \textbf{F1} & \textbf{Acc} & \textbf{Pre} & \textbf{Rec} & \textbf{F1} \\
    \midrule
    \multicolumn{13}{c}{\textit{\textbf{With Standard Answers}}} \\
    \midrule
    DeepSeek-V3-0324        & 80.0 & 68.6 & 93.6 & 79.2 & 87.1 & 78.7 & 93.6 & 85.5 & 95.1 & 94.3 & 93.6 & 93.9 \\
    DeepSeek-R1-0528        & 86.9 & 79.8 & 90.6 & 84.9 & 91.4 & 88.5 & 90.6 & 89.5 & 95.6 & 98.4 & 90.6 & 94.4 \\
    Qwen-Max-Preview        & 88.4 & 81.8 & 91.9 & 86.5 & 92.7 & 90.3 & 91.9 & 91.1 & 95.7 & 97.4 & 91.9 & 94.6 \\
    gpt-oss-120b(high)      & 88.9 & 82.6 & 92.1 & 87.1 & 93.2 & 91.2 & 92.1 & 91.7 & 96.3 & 98.7 & 92.1 & 95.3 \\
    Qwen2.5-Math-PRM-72B & 83.4 & 72.6 & 95.1 & 82.3 & 89.7 & 82.3 & 95.1 & 88.2 & 95.6 & 94.2 & 95.1 & 94.6 \\
    DeepSeek-R1-Distill-32B & 78.3 & 67.5 & 89.7 & 77.0 & 85.4 & 77.8 & 89.7 & 83.3 & 92.2 & 91.0 & 89.7 & 90.3 \\
    OPV                     & 86.5 & 77.4 & 94.3 & 85.0 & 92.8 & 88.7 & 94.3 & 91.4 & 96.8 & 97.7 & 94.3 & 96.0 \\
    \midrule
    \multicolumn{13}{c}{\textit{\textbf{Without Standard Answers}}} \\
    \midrule
    DeepSeek-V3-0324        & 79.8 & 68.0 & 95.1 & 79.3 & 86.5 & 77.0 & 95.1 & 85.1 & 93.3 & 89.1 & 95.1 & 92.0 \\
    DeepSeek-R1-0528        & 87.7 & 79.5 & 93.8 & 86.1 & 92.3 & 88.0 & 93.8 & 90.8 & 96.5 & 97.4 & 93.8 & 95.6 \\
    Qwen-Max-Preview        & 89.5 & 83.1 & 93.1 & 87.8 & 93.0 & 90.0 & 93.1 & 91.5 & 96.2 & 97.4 & 93.1 & 95.2 \\
    gpt-oss-120b(high)      & 90.1 & 84.3 & 92.9 & 88.4 & 94.0 & 92.4 & 92.9 & 92.6 & 96.9 & 99.5 & 92.9 & 96.1 \\
    Qwen2.5-Math-PRM-72B & 81.7 & 71.0 & 92.9 & 80.5 & 87.5 & 79.7 & 92.9 & 85.8 & 92.8 & 89.8 & 92.9 & 91.3 \\
    DeepSeek-R1-Distill-32B & 83.7 & 72.1 & 97.8 & 83.0 & 90.7 & 82.5 & 97.8 & 89.5 & 95.1 & 90.8 & 97.8 & 94.2 \\
    OPV                     & 87.5 & 78.6 & 95.1 & 86.1 & 93.0 & 88.5 & 95.1 & 91.7 & 96.7 & 96.7 & 95.1 & 95.9 \\
    \bottomrule
    \end{tabular}%
  }
  \label{tab:math_results}%
\end{table}

\begin{table}[H]
  \centering
  \vspace{-1mm}
  \caption{
  Detailed evaluation results on OlympiadBench partition of ProcessBench.
  }
  \vspace{-1mm}
  \resizebox{0.9\linewidth}{!}{
    \begin{tabular}{lcccccccccccc}
    \toprule
    \multirow{2}{*}{\textbf{Model}} & \multicolumn{4}{c}{\textbf{Absolute}} & \multicolumn{4}{c}{\textbf{Approximate}} & \multicolumn{4}{c}{\textbf{Rough}} \\
    \cmidrule(lr){2-5} \cmidrule(lr){6-9} \cmidrule(lr){10-13}
      & \textbf{Acc} & \textbf{Pre} & \textbf{Rec} & \textbf{F1} & \textbf{Acc} & \textbf{Pre} & \textbf{Rec} & \textbf{F1} & \textbf{Acc} & \textbf{Pre} & \textbf{Rec} & \textbf{F1} \\
    \midrule
    \multicolumn{13}{c}{\textit{\textbf{With Standard Answers}}} \\
    \midrule
    DeepSeek-V3-0324        & 73.4 & 56.7 & 91.4 & 70.0 & 82.2 & 67.6 & 91.4 & 77.7 & 93.8 & 90.6 & 91.4 & 91.0 \\
    DeepSeek-R1-0528        & 81.8 & 68.1 & 87.5 & 76.6 & 89.1 & 81.7 & 87.5 & 84.5 & 95.2 & 98.3 & 87.5 & 92.6 \\
    Qwen-Max-Preview        & 80.7 & 66.4 & 87.2 & 75.4 & 87.6 & 78.6 & 87.2 & 82.7 & 94.6 & 96.7 & 87.2 & 91.7 \\
    gpt-oss-120b(high)      & 81.4 & 68.1 & 85.1 & 75.7 & 87.7 & 79.9 & 85.1 & 82.4 & 94.6 & 99.0 & 85.1 & 91.5 \\
    Qwen2.5-Math-PRM-72B & 75.6 & 59.2 & 90.8 & 71.7 & 83.1 & 69.2 & 90.8 & 78.5 & 92.7 & 88.2 & 90.8 & 89.4 \\
    DeepSeek-R1-Distill-32B & 65.8 & 49.8 & 80.1 & 61.4 & 75.2 & 60.2 & 80.1 & 68.7 & 88.1 & 84.1 & 80.1 & 82.0 \\
    OPV                     & 78.1 & 62.4 & 89.6 & 73.6 & 85.3 & 73.2 & 89.6 & 80.6 & 95.5 & 97.1 & 89.6 & 93.2 \\
    \midrule
    \multicolumn{13}{c}{\textit{\textbf{Without Standard Answers}}} \\
    \midrule
    DeepSeek-V3-0324        & 69.9 & 53.4 & 90.5 & 67.2 & 78.8 & 63.2 & 90.5 & 74.4 & 90.7 & 83.5 & 90.5 & 86.9 \\
    DeepSeek-R1-0528        & 81.9 & 67.2 & 91.4 & 77.4 & 89.0 & 79.3 & 91.4 & 84.9 & 95.7 & 95.9 & 91.4 & 93.6 \\
    Qwen-Max-Preview        & 82.1 & 68.9 & 86.3 & 76.6 & 88.6 & 81.2 & 86.3 & 83.7 & 94.6 & 97.6 & 86.3 & 91.6 \\
    gpt-oss-120b(high)      & 82.6 & 69.6 & 86.3 & 77.1 & 88.7 & 81.4 & 86.3 & 83.8 & 95.0 & 99.0 & 86.3 & 92.2 \\
    Qwen2.5-Math-PRM-72B & 74.2 & 57.9 & 88.4 & 70.0 & 81.1 & 66.7 & 88.4 & 76.1 & 89.0 & 80.9 & 88.4 & 84.5 \\
    DeepSeek-R1-Distill-32B & 76.4 & 59.3 & 97.3 & 73.7 & 84.2 & 69.0 & 97.3 & 80.7 & 94.4 & 87.7 & 97.3 & 92.2 \\
    OPV                     & 79.3 & 64.3 & 87.8 & 74.2 & 86.2 & 75.6 & 87.8 & 81.3 & 94.9 & 97.0 & 87.8 & 92.2 \\
    \bottomrule
    \end{tabular}%
  }
  \label{tab:olympiadbench_results}%
\end{table}

\begin{table}[H]
  \centering
  \vspace{-1mm}
  \caption{
  Detailed evaluation results on Omni-MATH partition of ProcessBench.
  }
  \vspace{-1mm}
  \resizebox{0.9\linewidth}{!}{
    \begin{tabular}{lcccccccccccc}
    \toprule
    \multirow{2}{*}{\textbf{Model}} & \multicolumn{4}{c}{\textbf{Absolute}} & \multicolumn{4}{c}{\textbf{Approximate}} & \multicolumn{4}{c}{\textbf{Rough}} \\
    \cmidrule(lr){2-5} \cmidrule(lr){6-9} \cmidrule(lr){10-13}
      & \textbf{Acc} & \textbf{Pre} & \textbf{Rec} & \textbf{F1} & \textbf{Acc} & \textbf{Pre} & \textbf{Rec} & \textbf{F1} & \textbf{Acc} & \textbf{Pre} & \textbf{Rec} & \textbf{F1} \\
    \midrule
    \multicolumn{13}{c}{\textit{\textbf{With Standard Answers}}} \\
    \midrule
    DeepSeek-V3-0324        & 64.5 & 39.5 & 89.2 & 54.8 & 77.6 & 52.1 & 89.2 & 65.7 & 92.8 & 82.4 & 89.2 & 85.7 \\
    DeepSeek-R1-0528        & 74.0 & 47.7 & 82.2 & 60.4 & 82.8 & 60.6 & 82.2 & 69.7 & 94.2 & 93.0 & 82.2 & 87.2 \\
    Qwen-Max-Preview        & 77.4 & 51.9 & 84.2 & 64.2 & 85.6 & 65.7 & 84.2 & 73.8 & 94.7 & 93.1 & 84.2 & 88.5 \\
    gpt-oss-120b(high)      & 76.9 & 51.3 & 83.4 & 63.5 & 85.8 & 66.3 & 83.4 & 73.9 & 95.4 & 97.1 & 83.4 & 89.7 \\
    Qwen2.5-Math-PRM-72B & 67.8 & 42.2 & 91.3 & 57.7 & 78.1 & 52.6 & 91.3 & 66.8 & 91.0 & 76.1 & 91.3 & 83.0 \\
    DeepSeek-R1-Distill-32B & 58.0 & 34.6 & 83.8 & 49.0 & 71.6 & 45.2 & 83.8 & 58.7 & 88.3 & 72.1 & 83.8 & 77.5 \\
    OPV                     & 71.2 & 44.8 & 84.6 & 58.6 & 82.0 & 58.8 & 84.6 & 69.4 & 95.4 & 95.8 & 84.6 & 89.9 \\
    \midrule
    \multicolumn{13}{c}{\textit{\textbf{Without Standard Answers}}} \\
    \midrule
    DeepSeek-V3-0324        & 61.5 & 37.4 & 88.4 & 52.5 & 74.5 & 48.4 & 88.4 & 62.6 & 90.7 & 76.6 & 88.4 & 82.1 \\
    DeepSeek-R1-0528        & 76.7 & 51.0 & 88.0 & 64.5 & 84.5 & 62.7 & 88.0 & 73.2 & 95.2 & 91.8 & 88.0 & 89.8 \\
    Qwen-Max-Preview        & 78.6 & 53.7 & 80.9 & 64.6 & 85.7 & 66.8 & 80.9 & 73.2 & 94.2 & 94.2 & 80.9 & 87.1 \\
    gpt-oss-120b(high)      & 78.0 & 52.8 & 83.0 & 64.5 & 85.9 & 66.7 & 83.0 & 73.9 & 95.4 & 97.6 & 83.0 & 89.7 \\
    Qwen2.5-Math-PRM-72B & 69.9 & 43.8 & 88.4 & 58.6 & 79.7 & 54.9 & 88.4 & 67.7 & 90.4 & 75.8 & 88.4 & 81.6 \\
    DeepSeek-R1-Distill-32B & 69.0 & 43.6 & 97.1 & 60.2 & 78.9 & 53.4 & 97.1 & 68.9 & 92.9 & 78.5 & 97.1 & 86.8 \\
    OPV                     & 72.5 & 46.1 & 83.0 & 59.3 & 82.7 & 60.2 & 83.0 & 69.8 & 95.2 & 96.6 & 83.0 & 89.3 \\
    \bottomrule
    \end{tabular}%
  }
  \label{tab:omnimath_results}%
\end{table}


\begin{table}[H]
  \centering
  \vspace{2mm}
  \caption{
  Detailed evaluation results on K-12 partition of \thisbench.
  }
  \vspace{-1mm}
  \resizebox{0.9\linewidth}{!}{
    \begin{tabular}{lcccccccccccc}
    \toprule
    \multirow{2}{*}{\textbf{Model}} & \multicolumn{4}{c}{\textbf{Absolute}} & \multicolumn{4}{c}{\textbf{Approximate}} & \multicolumn{4}{c}{\textbf{Rough}} \\
    \cmidrule(lr){2-5} \cmidrule(lr){6-9} \cmidrule(lr){10-13}
      & \textbf{Acc} & \textbf{Pre} & \textbf{Rec} & \textbf{F1} & \textbf{Acc} & \textbf{Pre} & \textbf{Rec} & \textbf{F1} & \textbf{Acc} & \textbf{Pre} & \textbf{Rec} & \textbf{F1} \\
    \midrule
    \multicolumn{13}{c}{\textit{\textbf{With Standard Answers}}} \\
    \midrule
    DeepSeek-V3-0324 & 75.7 & 80.3 & 78.3 & 79.3 & 77.7 & 83.2 & 78.3 & 80.7 & 81.2 & 88.7 & 78.3 & 83.2 \\
        DeepSeek-R1-0528        & 86.6 & 85.4 & 93.3 & 89.2 & 87.1 & 86.1 & 93.3 & 89.5 & 88.6 & 88.1 & 93.3 & 90.6 \\
    Qwen-Max-Preview        & 87.6 & 87.9 & 91.6 & 89.7 & 88.6 & 89.3 & 91.6 & 90.5 & 90.0 & 91.6 & 91.6 & 91.6 \\
    gpt-oss-120b(high)      & 82.1 & 79.9 & 93.3 & 86.1 & 84.6 & 82.8 & 93.3 & 87.8 & 87.6 & 86.7 & 93.3 & 89.9 \\
    Qwen2.5-Math-PRM-72B & 60.4 & 61.4 & 90.0 & 73.0 & 61.4 & 62.1 & 90.0 & 73.5 & 68.8 & 67.9 & 90.0 & 77.4 \\
    DeepSeek-R1-Distill-32B & 75.1 & 74.5 & 88.2 & 80.8 & 77.1 & 76.6 & 88.2 & 82.0 & 80.6 & 80.8 & 88.2 & 84.3 \\
    OPV-Stage1              & 76.6 & 72.2 & 98.3 & 83.3 & 78.6 & 74.1 & 98.3 & 84.5 & 81.1 & 76.5 & 98.3 & 86.0 \\
    OPV-Stage2              & 86.6 & 82.4 & 98.3 & 89.7 & 87.6 & 83.6 & 98.3 & 90.4 & 90.5 & 87.3 & 98.3 & 92.5 \\
    OPV-Stage3-w/o RL       & 87.6 & 88.5 & 90.8 & 89.6 & 88.6 & 90.0 & 90.8 & 90.4 & 90.0 & 92.3 & 90.8 & 91.5 \\
    OPV                     & 86.1 & 89.6 & 86.6 & 88.0 & 87.1 & 91.2 & 86.6 & 88.8 & 88.1 & 92.8 & 86.6 & 89.6 \\

    \midrule
    \multicolumn{13}{c}{\textit{\textbf{Without Standard Answers}}} \\
    \midrule
    DeepSeek-V3-0324 & 70.3 & 71.7 & 82.5 & 76.7 & 71.8 & 73.3 & 82.5 & 77.7 & 76.2 & 78.6 & 82.5 & 80.5 \\
        DeepSeek-R1-0528        & 83.1 & 80.1 & 95.0 & 86.9 & 83.1 & 80.1 & 95.0 & 86.9 & 85.1 & 82.5 & 95.0 & 88.3 \\
    Qwen-Max-Preview        & 84.6 & 83.3 & 92.4 & 87.7 & 86.1 & 85.3 & 92.4 & 88.7 & 87.6 & 87.3 & 92.4 & 89.8 \\
    gpt-oss-120b(high)      & 79.0 & 77.7 & 90.8 & 83.7 & 81.0 & 80.0 & 90.8 & 85.0 & 85.0 & 85.0 & 90.8 & 87.8 \\
    Qwen2.5-Math-PRM-72B & 61.9 & 62.9 & 87.5 & 73.2 & 62.4 & 63.3 & 87.5 & 73.4 & 64.4 & 64.8 & 87.5 & 74.5 \\
    DeepSeek-R1-Distill-32B & 73.6 & 70.6 & 95.0 & 81.0 & 74.6 & 71.5 & 95.0 & 81.6 & 77.1 & 73.9 & 95.0 & 83.1 \\
    OPV-Stage1              & 72.6 & 69.3 & 96.6 & 80.7 & 73.1 & 69.7 & 96.6 & 81.0 & 74.6 & 71.0 & 96.6 & 81.9 \\
    OPV-Stage2              & 83.1 & 80.1 & 95.0 & 86.9 & 83.6 & 80.7 & 95.0 & 87.3 & 87.1 & 85.0 & 95.0 & 89.7 \\
    OPV-Stage3-w/o RL       & 84.6 & 84.4 & 90.8 & 87.5 & 85.1 & 85.0 & 90.8 & 87.8 & 87.1 & 87.8 & 90.8 & 89.3 \\
    OPV                     & 87.6 & 86.7 & 93.3 & 89.9 & 88.1 & 87.4 & 93.3 & 90.2 & 90.5 & 91.0 & 93.3 & 92.1 \\

    \bottomrule
    \end{tabular}%
  }
  \label{tab:k12_results}%
\end{table}

\begin{table}[H]
  \centering
  \vspace{2mm}
  \caption{
  Detailed evaluation results on High School Competition partition of \thisbench.
  }
  \vspace{-1mm}
  \resizebox{0.9\linewidth}{!}{
    \begin{tabular}{lcccccccccccc}
    \toprule
    \multirow{2}{*}{\textbf{Model}} & \multicolumn{4}{c}{\textbf{Absolute}} & \multicolumn{4}{c}{\textbf{Approximate}} & \multicolumn{4}{c}{\textbf{Rough}} \\
    \cmidrule(lr){2-5} \cmidrule(lr){6-9} \cmidrule(lr){10-13}
      & \textbf{Acc} & \textbf{Pre} & \textbf{Rec} & \textbf{F1} & \textbf{Acc} & \textbf{Pre} & \textbf{Rec} & \textbf{F1} & \textbf{Acc} & \textbf{Pre} & \textbf{Rec} & \textbf{F1} \\
    \midrule
    \multicolumn{13}{c}{\textit{\textbf{With Standard Answers}}} \\
    \midrule
    DeepSeek-V3-0324 & 72.3 & 64.6 & 88.1 & 74.6 & 78.4 & 71.6 & 88.1 & 79.0 & 85.3 & 81.5 & 88.1 & 84.7 \\
        DeepSeek-R1-0528        & 76.0 & 69.4 & 93.5 & 79.7 & 82.3 & 76.4 & 93.5 & 84.1 & 88.9 & 85.7 & 93.5 & 89.4 \\
    Qwen-Max-Preview        & 73.4 & 67.5 & 90.8 & 77.4 & 79.4 & 74.0 & 90.8 & 81.6 & 87.3 & 84.9 & 90.8 & 87.7 \\
    gpt-oss-120b(high)      & 68.9 & 64.2 & 86.1 & 73.5 & 78.6 & 75.1 & 86.1 & 80.2 & 86.3 & 86.5 & 86.1 & 86.3 \\
    Qwen2.5-Math-PRM-72B & 49.5 & 47.0 & 97.6 & 63.4 & 51.3 & 47.9 & 97.6 & 64.3 & 56.6 & 50.9 & 97.6 & 66.9 \\
    DeepSeek-R1-Distill-32B & 73.4 & 66.0 & 97.0 & 78.6 & 79.1 & 71.6 & 97.0 & 82.4 & 83.1 & 76.0 & 97.0 & 85.3 \\
    OPV-Stage1              & 71.8 & 64.5 & 97.5 & 77.6 & 76.3 & 68.5 & 97.5 & 80.5 & 80.1 & 72.5 & 97.5 & 83.1 \\
    OPV-Stage2              & 75.8 & 69.0 & 94.0 & 79.6 & 80.8 & 74.4 & 94.0 & 83.1 & 87.0 & 82.5 & 94.0 & 87.9 \\
    OPV-Stage3-w/o RL       & 79.9 & 78.6 & 82.3 & 80.4 & 85.5 & 88.0 & 82.3 & 85.1 & 90.3 & 97.9 & 82.3 & 89.5 \\
    OPV                     & 83.0 & 81.7 & 85.3 & 83.5 & 87.9 & 90.0 & 85.3 & 87.6 & 91.3 & 96.9 & 85.3 & 90.7 \\

    \midrule
    \multicolumn{13}{c}{\textit{\textbf{Without Standard Answers}}} \\
    \midrule
    DeepSeek-V3-0324 & 66.5 & 58.6 & 90.7 & 71.2 & 73.1 & 64.6 & 90.7 & 75.4 & 81.2 & 74.0 & 90.7 & 81.5 \\
        DeepSeek-R1-0528        & 67.0 & 61.4 & 92.8 & 73.9 & 74.5 & 68.1 & 92.8 & 78.5 & 83.3 & 78.0 & 92.8 & 84.8 \\
    Qwen-Max-Preview        & 69.4 & 63.8 & 90.3 & 74.8 & 77.6 & 72.2 & 90.3 & 80.2 & 86.4 & 83.8 & 90.3 & 87.0 \\
    gpt-oss-120b(high)      & 67.1 & 62.7 & 85.3 & 72.3 & 76.9 & 73.1 & 85.3 & 78.8 & 84.9 & 84.7 & 85.3 & 85.0 \\
    Qwen2.5-Math-PRM-72B & 49.5 & 47.0 & 97.3 & 63.4 & 51.1 & 47.8 & 97.3 & 64.1 & 54.4 & 49.6 & 97.3 & 65.7 \\
    DeepSeek-R1-Distill-32B & 64.6 & 59.0 & 97.5 & 73.5 & 68.4 & 61.7 & 97.5 & 75.6 & 73.1 & 65.7 & 97.5 & 78.5 \\
    OPV-Stage1              & 65.8 & 59.9 & 96.3 & 73.9 & 70.8 & 63.9 & 96.3 & 76.8 & 76.1 & 68.7 & 96.3 & 80.2 \\
    OPV-Stage2              & 69.8 & 63.6 & 93.3 & 75.6 & 77.8 & 71.3 & 93.3 & 80.8 & 85.4 & 80.7 & 93.3 & 86.5 \\
    OPV-Stage3-w/o RL       & 76.4 & 72.1 & 86.3 & 78.6 & 82.4 & 80.1 & 86.3 & 83.1 & 89.0 & 91.3 & 86.3 & 88.8 \\
    OPV                     & 78.3 & 72.6 & 91.0 & 80.8 & 85.4 & 81.9 & 91.0 & 86.2 & 90.8 & 90.6 & 91.0 & 90.8 \\

    \bottomrule
    \end{tabular}%
  }
  \label{tab:highschool_results}%
\end{table}

\begin{table}[H]
  \centering
  \vspace{2mm}
  \caption{
  Detailed evaluation results on Undergraduate partition of \thisbench.
  }
  \vspace{-1mm}
  \resizebox{0.9\linewidth}{!}{
    \begin{tabular}{lcccccccccccc}
    \toprule
    \multirow{2}{*}{\textbf{Model}} & \multicolumn{4}{c}{\textbf{Absolute}} & \multicolumn{4}{c}{\textbf{Approximate}} & \multicolumn{4}{c}{\textbf{Rough}} \\
    \cmidrule(lr){2-5} \cmidrule(lr){6-9} \cmidrule(lr){10-13}
      & \textbf{Acc} & \textbf{Pre} & \textbf{Rec} & \textbf{F1} & \textbf{Acc} & \textbf{Pre} & \textbf{Rec} & \textbf{F1} & \textbf{Acc} & \textbf{Pre} & \textbf{Rec} & \textbf{F1} \\
    \midrule
    \multicolumn{13}{c}{\textit{\textbf{With Standard Answers}}} \\
    \midrule
    DeepSeek-V3-0324 & 62.2 & 59.1 & 78.8 & 67.5 & 66.6 & 63.3 & 78.8 & 70.2 & 71.5 & 68.7 & 78.8 & 73.4 \\
        DeepSeek-R1-0528        & 58.3 & 57.0 & 68.9 & 62.4 & 63.6 & 62.5 & 68.9 & 65.5 & 69.9 & 70.5 & 68.9 & 69.7 \\
    Qwen-Max-Preview        & 58.7 & 56.9 & 73.3 & 64.1 & 64.2 & 62.1 & 73.3 & 67.3 & 70.0 & 68.9 & 73.3 & 71.1 \\
    gpt-oss-120b(high)      & 52.8 & 52.4 & 67.4 & 58.9 & 62.0 & 61.1 & 67.4 & 64.1 & 70.7 & 72.5 & 67.4 & 69.8 \\
    Qwen2.5-Math-PRM-72B & 56.2 & 53.5 & 95.2 & 68.5 & 57.6 & 54.3 & 95.2 & 69.2 & 62.0 & 57.2 & 95.2 & 71.5 \\
    DeepSeek-R1-Distill-32B & 67.9 & 62.3 & 91.5 & 74.1 & 71.7 & 65.7 & 91.5 & 76.5 & 75.4 & 69.4 & 91.5 & 78.9 \\
    OPV-Stage1              & 65.8 & 60.3 & 93.3 & 73.3 & 69.4 & 63.2 & 93.3 & 75.4 & 72.7 & 66.2 & 93.3 & 77.5 \\
    OPV-Stage2              & 66.5 & 62.7 & 82.2 & 71.1 & 71.4 & 67.8 & 82.2 & 74.3 & 75.7 & 72.9 & 82.2 & 77.2 \\
    OPV-Stage3-w/o RL       & 72.2 & 74.2 & 68.6 & 71.3 & 76.2 & 81.0 & 68.6 & 74.3 & 81.0 & 91.5 & 68.6 & 78.4 \\
    OPV                     & 75.1 & 76.6 & 72.5 & 74.5 & 79.6 & 84.7 & 72.5 & 78.1 & 84.3 & 95.1 & 72.5 & 82.3 \\

    \midrule
    \multicolumn{13}{c}{\textit{\textbf{Without Standard Answers}}} \\
    \midrule
    DeepSeek-V3-0324 & 53.3 & 52.2 & 79.2 & 62.9 & 57.9 & 55.5 & 79.2 & 65.3 & 63.9 & 60.6 & 79.2 & 68.7 \\
        DeepSeek-R1-0528        & 46.4 & 47.5 & 64.3 & 54.7 & 52.9 & 52.5 & 64.3 & 57.8 & 62.3 & 62.0 & 64.3 & 63.1 \\
    Qwen-Max-Preview        & 51.9 & 51.6 & 68.3 & 58.8 & 57.1 & 56.0 & 68.3 & 61.5 & 65.8 & 65.3 & 68.3 & 66.7 \\
    gpt-oss-120b(high)      & 48.9 & 49.3 & 63.0 & 55.3 & 58.2 & 57.7 & 63.0 & 60.2 & 68.2 & 70.5 & 63.0 & 66.5 \\
    Qwen2.5-Math-PRM-72B & 55.2 & 52.9 & 94.0 & 67.7 & 57.1 & 54.1 & 94.0 & 68.7 & 60.4 & 56.2 & 94.0 & 70.4 \\
    DeepSeek-R1-Distill-32B & 58.0 & 55.1 & 88.9 & 68.0 & 61.3 & 57.4 & 88.9 & 69.8 & 65.7 & 60.9 & 88.9 & 72.3 \\
    OPV-Stage1              & 56.0 & 53.8 & 86.3 & 66.3 & 59.7 & 56.4 & 86.3 & 68.2 & 64.9 & 60.6 & 86.3 & 71.2 \\
    OPV-Stage2              & 58.2 & 56.4 & 73.7 & 63.9 & 63.2 & 61.1 & 73.7 & 66.8 & 71.2 & 70.3 & 73.7 & 71.9 \\
    OPV-Stage3-w/o RL       & 61.4 & 60.8 & 65.1 & 62.9 & 67.4 & 68.4 & 65.1 & 66.7 & 73.1 & 77.6 & 65.1 & 70.8 \\
    OPV                     & 65.4 & 63.7 & 72.5 & 67.8 & 71.9 & 71.8 & 72.5 & 72.2 & 77.0 & 79.8 & 72.5 & 76.0 \\

    \bottomrule
    \end{tabular}%
  }
  \label{tab:undergraduate_results}%
\end{table}

\section{Case Studies}
Here we list interesting cases containing process errors that were found by \thiswork\ across different datasets.

\begin{figure*}[t] 
\begin{AIbox}{}
{\bf Question:} \\
{
Let \( a, b, c > 0 \) such that \( a + b \leq c \). Find the minimum value of 
\( P = \left( a^4 + b^4 + c^4 \right) \left( \frac{1}{a^4} + \frac{1}{b^4} + \frac{1}{c^4} \right). \)
} \\\\
{\bf Model Answer from AM-DeepSeek-R1-0528-Distilled:} \\
{
Given positive real numbers \(a\), \(b\), and \(c\) with the constraint \(a + b \leq c\), the goal is to minimize the expression
\[
P = \left( a^4 + b^4 + c^4 \right) \left( \frac{1}{a^4} + \frac{1}{b^4} + \frac{1}{c^4} \right).
\]

\textcolor{red}{The expression \(P\) is symmetric in \(a\) and \(b\), and the constraint is also symmetric in \(a\) and \(b\). Therefore, the minimum likely occurs when \(a = b\). Assuming \(a = b\), the constraint becomes \(2a \leq c\). Let \(t = c / a\), so \(t \geq 2\).} Then,
\[
P = \left( a^4 + a^4 + (t a)^4 \right) \left( \frac{1}{a^4} + \frac{1}{a^4} + \frac{1}{(t a)^4} \right) = (2a^4 + t^4 a^4) \left( \frac{2}{a^4} + \frac{1}{t^4 a^4} \right).
\]
The \(a^4\) terms cancel, giving
\[
P = (2 + t^4) \left( 2 + \frac{1}{t^4} \right) = 4 + 2t^4 + \frac{2}{t^4} + 1 = 5 + 2t^4 + \frac{2}{t^4}.
\]
Substituting \(u = t^4\) (so \(u \geq 16\)),\(P = 5 + 2u + \frac{2}{u}\).
The derivative is \(P'(u) = 2 - \frac{2}{u^2}\), which is positive for \(u > 1\). Since \(u \geq 16 > 1\), \(P(u)\) is increasing for \(u \geq 16\), and the minimum occurs at \(u = 16\) (\ie, \(t = 2\)). At this point,
\[
P = 5 + 2(16) + \frac{2}{16} = 5 + 32 + 0.125 = 37.125 = \frac{297}{8}.
\]
This value is achieved when \(a = b\) and \(c = 2a\), satisfying \(a + b = c\) (equality in the constraint).

\textcolor{red}{To confirm this is the global minimum, consider cases where \(a \neq b\) or \(a + b < c\).} When \(a + b < c\), setting \(a = b = 1\) and \(c > 2\) gives \(P > 297/8\). For \(a \neq b\) with \(a + b = c\), define \(r = b / a > 0\), so
\[
P = \left[1 + r^4 + (1 + r)^4\right] \left[1 + \frac{1}{r^4} + \frac{1}{(1 + r)^4}\right].
\]
At \(r = 1\), \(P = 297/8\). The derivative at \(r = 1\) is zero, and \(P(r) > 297/8\) for \(r \neq 1\). As \(r \to 0^+\) or \(r \to \infty\), \(P \to \infty\). Thus, under \(a + b = c\), the minimum is \(297/8\). When \(a + b < c\), \(P > 297/8\), so the minimum is achieved when equality holds in the constraint.

Thus, the minimum value of \(P\) is \(\frac{297}{8}\).
} \\\\
{\bf Judgments of \thiswork:} \\
{
The first error occurs in Step 1 where the student assumes \(a = b\) based solely on symmetry without rigorous justification. While the problem is symmetric in \(a\) and \(b\), the minimum might not necessarily occur at \(a = b\) without proof. Although the student later verifies this assumption through additional analysis, the initial step lacks a formal argument (\eg, using inequalities, convexity, or perturbation methods) to confirm that deviations from \(a = b\) would increase \(P\). This makes Step 1 logically incomplete and thus incorrect.
}
\end{AIbox} 
\caption{\modified{\textbf{A case study sampled from AM-DeepSeek-R1-0528-Distilled dataset where the reasoning process is defective despite the final answer being correct.} The rigor of the inequality is not proven in the model's proposed answer, which is identified by \thiswork. In the generations, we use ref color to present the \textcolor{red}{defective content}.}}

\label{fig: case study}
\end{figure*}



\end{document}